\documentclass[10pt,twocolumn,letterpaper]{article}

\usepackage[pagenumbers]{cvpr} % To force page numbers, e.g. for an arXiv version

\usepackage[dvipsnames]{xcolor}

\usepackage{soul}
\definecolor{cvprblue}{rgb}{0.21,0.49,0.74}
\usepackage[pagebackref,breaklinks,colorlinks,citecolor=cvprblue]{hyperref}

\usepackage{tabularx}
\usepackage{multirow}
\usepackage{arydshln}
\usepackage{pifont}% http://ctan.org/pkg/pifont
\newcommand{\cmark}{\ding{51}}%
\newcommand{\xmark}{\ding{55}}%
\usepackage{makecell}
\usepackage[normalem]{ulem}

\def\paperID{2496} % *** Enter the Paper ID here
\def\confName{CVPR}
\def\confYear{2024}

\title{
SI-MIL: Taming Deep MIL for Self-Interpretability in Gigapixel Histopathology
}

\newcommand*\samethanks[1][\value{footnote}]{\footnotemark[#1]}

\author{Saarthak Kapse$^{1}$\thanks{These authors contributed equally to this paper.} , Pushpak Pati$^{4}$\samethanks[1] , Srijan Das$^2$, Jingwei Zhang$^1$, Chao Chen$^1$, Maria Vakalopoulou$^3$, \\ Joel Saltz$^1$,  Dimitris Samaras$^1$, Rajarsi R. Gupta$^1$, Prateek Prasanna$^1$ \\ \\
$^1$Stony Brook University, USA\\$^2$UNC Charlotte, USA \\ $^3$CentraleSupélec, University of Paris-Saclay, France\\ $^4$Independent Researcher
}

\begin{document}
\maketitle
\begin{abstract}

% \cc{Need to be polished according to current intro.}

Introducing interpretability and reasoning into Multiple Instance Learning (MIL) methods for Whole Slide Image (WSI) analysis is challenging, given the complexity of gigapixel slides. Traditionally, MIL interpretability is limited to identifying salient regions deemed pertinent for downstream tasks, offering little insight to the end-user (pathologist) regarding the rationale behind these selections. To address this, we propose Self-Interpretable MIL (SI-MIL), a method intrinsically designed for interpretability from the very outset. SI-MIL employs a deep MIL framework to guide an interpretable branch grounded on handcrafted pathological features, facilitating linear predictions. Beyond identifying salient regions, SI-MIL uniquely provides feature-level interpretations rooted in pathological insights for WSIs. Notably, SI-MIL, with its linear prediction constraints, challenges the prevalent myth of an inevitable trade-off between model interpretability and performance, demonstrating competitive results compared to state-of-the-art methods on WSI-level prediction tasks across three cancer types. In addition, we thoroughly benchmark the local- and global-interpretability of SI-MIL in terms of statistical analysis, a domain expert study, and desiderata of interpretability, namely, user-friendliness and faithfulness. 
%through a domain expert study, we verify the faithfulness of the patch-feature importance analysis enabled by SI-MIL.
\begingroup
\renewcommand\thefootnote{}\footnote{
    \hspace{-1em}Code and Dataset is available at: \href{https://github.com/bmi-imaginelab/SI-MIL/tree/main}{github.com/bmi-imaginelab/SI-MIL}
    % \url{https://github.com/bmi-imaginelab/SI-MIL/tree/main}
}
\addtocounter{footnote}{-1}% Optional: Reset the footnote counter
\endgroup
\vspace{-0.3cm}

% We introduce a Self-Interpretable Multiple Instance Learning (SI-MIL) method for computational pathology to address the critical need for transparency in gigapixel whole slide image (WSI) analysis. SI-MIL successfully overcomes the traditional accuracy-interpretability tradeoff of using  interpretable pathological features instead of deep learning features by integrating them in a novel dual-branch network. This approach not only maintains competitive predictive power but also provides clear interpretability at both patient and dataset levels. We demonstrate that our method extends beyond the typical capability of MIL systems, which often only identify key regions in a WSI. Our approach not only highlights these important areas but since SI-MIL is grounded by hand-crafted pathological feature, it can also elucidates the specific features within these regions that drive the prediction. This provides a holistically interpretable framework that can influence how pathologists interact with automated systems. Notably, SI-MIL represents the first method to be interpretable-by-design at both patch and feature level while capable of scaling to gigapixel WSIs.

\end{abstract}
    
\section{Introduction}
\label{sec:intro}

In the last decade, advancements in deep learning techniques, especially Multiple Instance Learning (MIL) algorithms~\cite{ilse2018attention, lu2021data, shao2021transmil}, have dramatically revolutionized computational pathology, which has transitioned from analyzing local regions-of-interest~\cite{madabhushi2016image} to gigapixel whole slide images (WSIs). A standard MIL workflow takes in feature representations of patches from a WSI, embedded via a deep neural network, and aggregates them to define a slide-level representation adept for a downstream task. While these deep neural network-reliant workflows have resulted in high performance, they often lack pathologist-friendly interpretability and reasoning in their predictions~\cite{rymarczyk2022protomil}, which is crucial for building trust in routine clinical workflows and defining reliability and accountability of AI algorithms~\cite{salahuddin2022transparency, border2022growing,graziani2023global}, particularly in clinical contexts.

\fboxsep=1pt%padding thickness
\fboxrule=1pt%border thickness
\definecolor{darkgreen}{rgb}{0.0, 0.8, 0.0}

\begin{figure}[!t]
\centering
    \includegraphics[width=1\linewidth]{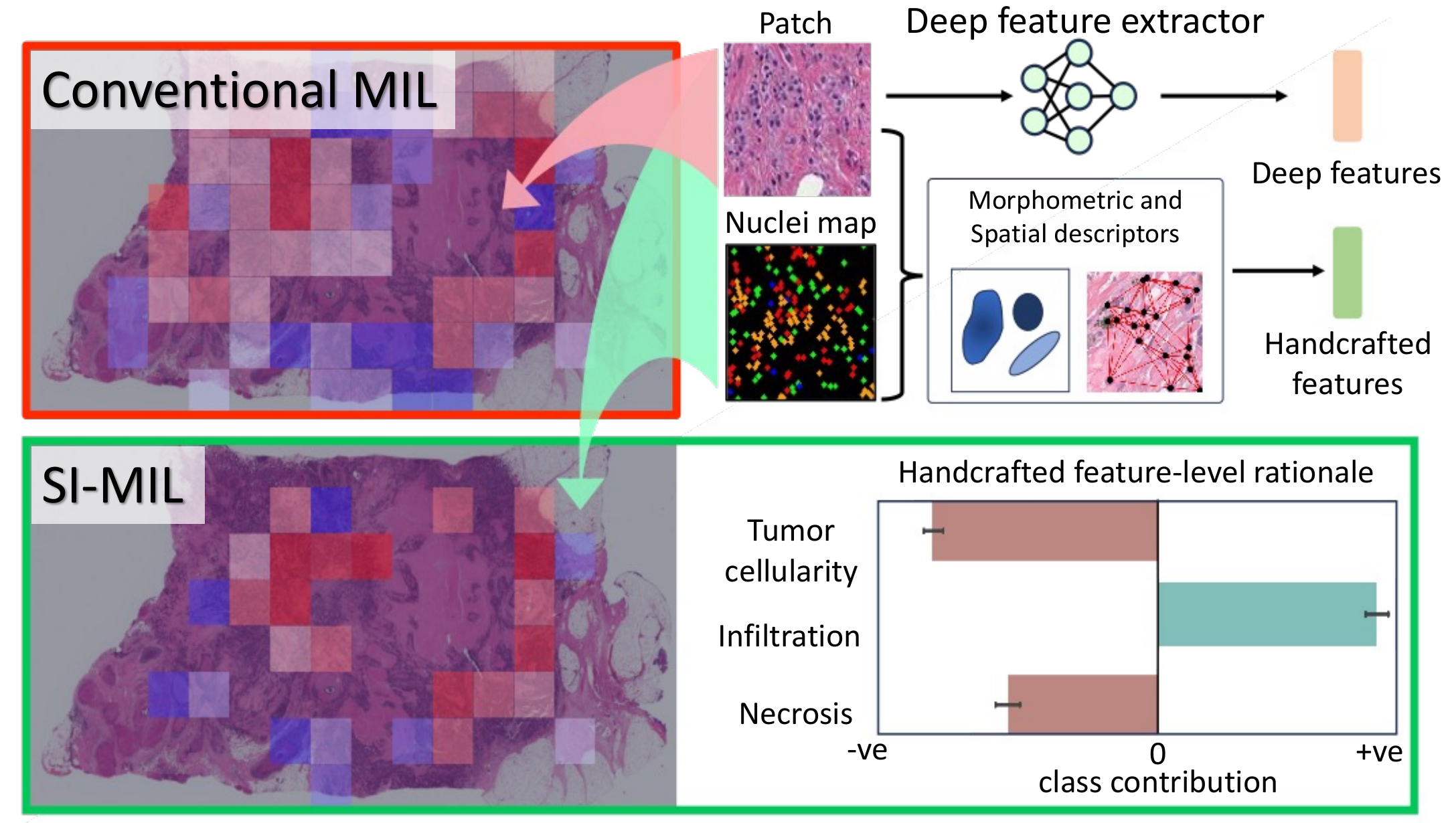} % SI-MIL teaser small realimg
    \caption{
    Unlike conventional \fcolorbox{red}{white}{MIL}, \fcolorbox{darkgreen}{white}{SI-MIL} co-learns from deep and handcrafted features (referred to as PathExpert features). While both MILs offer patch-level interpretability, only ours provides PathExpert feature-level rationale for WSI predictions. The attention maps in SI-MIL are grounded on geometrically and physically-interpretable descriptors.
    }
    \label{fig:teaser}
    \vspace{-0.7cm}
\end{figure}
%  draw 2 arrows from features to show which one is for what. 

In computational pathology, efforts to achieve WSI-level interpretability have predominantly focused on two directions: (1) identifying salient regions in a WSI, and (2) employing post-hoc techniques to elucidate the underlying patterns in salient regions. The first approach, employed by traditional MIL, includes techniques such as visualization of attention maps~\cite{ilse2018attention, lu2021data, shao2021transmil, zhang2023prompt, sureka2020visualization, zhang2021joint} and post-hoc gradient-based saliency~\cite{zheng2022graph, pati2023weakly}, which highlight image patches that influence the model prediction. 
% Though useful, these techniques may not necessarily provide a holistic understanding of why the model made a specific decision.
% , which confines to only a local perspective of the model
% Furthermore, the resulting visualizations might be challenging to interpret for the experts~\cite{arrieta2020explainable}, owning to the lack of expert-friendly feature grounding. 
These techniques, though useful, may not offer a complete understanding of the model's decisions and can result in visualizations that are hard for experts to interpret due to a lack of user-friendly feature grounding~\cite{arrieta2020explainable}.
The latter approach involves extracting interpretable handcrafted features (henceforth referred to as PathExpert features) from the MIL-identified salient patches and then conducting statistical analyses to find correlations between these features and the WSI ground truths in a post-hoc manner~\cite{mckenzie2022interpretable, chen2022pan, fremond2023interpretable}.
% However, as highlighted in the seminal work~\cite{rudin2019stop}, post-hoc methods can suffer from impaired reasoning due to their independence from the model’s actual computations. In the context of the above post-hoc methods~\cite{mckenzie2022interpretable, chen2022pan, fremond2023interpretable}, this independence is evident from the disconnect between the features used for MIL training, i.e., deep features, and those analyzed by statistical methods, i.e., PathExpert features. 
However, there is a clear disconnect between the deep features used for MIL training and the PathExpert features.
Using post-hoc PathExpert features to explain deep models can be sub-optimal~\cite{rudin2019stop}.
% In practice, this disconnect often leads to sub-optimal explanation; 
Furthermore, patches with high attention may be crucial for deep feature space, but may not be optimal in PathExpert feature space, thus compromising interpretability. 

To truly interpret a prediction model, it seems inevitable to bring interpretable features into training. A natural idea is to directly train MILs using these features, followed by statistical analysis of features from highly attended patches.
However, this will not exploit the full potential of deep learning, as our analysis in Section \ref{sec:results} will reveal. 
This brings us to the main question: \emph{Can we really achieve inherent interpretability without compromising model performance?}

The answer is yes.
In this paper, we  provide \textit{the first WSI solution with both inherent interpretability and strong prediction power}. Our key observation is that a highly accurate deep model is not unique; there can be many optimal or close-to-optimal deep models for a dataset/task, due to over-parameterization~\cite{nguyen2017loss,kawaguchi2016deep,laurent2018deep}.
Therefore, we hypothesize that we can alter the learning procedure and find an alternative model with desired interpretability and still be powerful in prediction.
In particular, we propose to pair a deep MIL model with an interpretable model grounded on PathExpert features during training. Through co-learning, the MIL retains its predictive power. Meanwhile, it is sufficiently ``tamed'' by the co-learned interpretable model, which renders interpretability. 
% And the interpretable model can provide well-grounded explanation to the decisions.
As shown in Fig.~\ref{fig:teaser}, the tamed deep MIL model has a different attention map from the standard MIL. It is attending to patches which can also be discriminated by the companion interpretable model. 
% In fact, our end product will be the co-learned interpretable model. %\emph{To the best of our knowledge, it is the first while-box model with as good prediction power as a deep MIL model.} \cc{Not 100\% sure this is true. After all, the attention is not white-box.}

%Given the effectiveness of deep features in MIL literature, we hypothesize that intelligently combining deep and PathExpert features within a new MIL could not only build an interpretable model but also  narrow their performance gap with their non-interpretable counterparts. 
%% A promising strategy involves developing a new method that simultaneously trains MIL with both sets of features.
%% A joint optimization of MIL with both the features could address the disconnect between the use of deep features for MIL and reliance on PathExpert features for interpretability in current post-hoc methods.  
%Therefore, a joint optimization of MIL with both the features might result in the optimal selection of patches since it also incorporates the impact PathExpert features as illustrated in fig.~\ref{fig:teaser}. 

% In this paper, we propose Self-Interpretable MIL (SI-MIL) designed with three primary goals: (1) strong predictive capability, (2) interpretability embedded into the learning process from the very outset, and (3) scalability to gigapixel-scale pathology images.
Our method, Self-Interpretable MIL (SI-MIL), is a dual branch network, consisting of a conventional MIL and a novel \textit{Self-Interpretable} (SI) branch. 
% The MIL branch leverages deep features in a routine manner and identifies salient patches in a WSI. 
% The MIL branch exploits deep features' robust discriminative power to guide the patch selection for the \textit{Self-Interpretable} branch. 
The MIL exploits deep features' discriminative power to guide the SI branch. 
% Grounded on PathExpert features from the salient patches, the Self-Interpretable branch provides a linear prediction.
Grounded on PathExpert features, the SI branch then provides a linear prediction.
% The \textit{Self-Interpretable} branch leverages PathExpert features from these salient patches; it then employs a feature attention module and a linear predictor to provide a prediction. 
A differentiable Top-$K$ operator for selecting patches, connects the two branches and enables end-to-end co-learning. 
% This enables both the deep and PathExpert feature sets to mutually optimize the selection of salient patches.
% \pp{Where do we include the 1-2 sentences on why we call our model SI?}
To highlight, SI-MIL is inherently interpretable~\cite{rudin2019stop, barbiero2023interpretable} due to the linear mapping between the PathExpert features and the output predictions. Therefore, it can reflect the impact of each feature on the output, thus providing a feature-level rationale, as shown in  Fig.~\ref{fig:teaser}.
Also, by leveraging the potential of a deep feature extractor, state-of-the-art MIL, and geometrically and physically- interpretable PathExpert features, SI-MIL counters the well-known myth of unavoidable model interpretability and performance trade-off~\cite{arrieta2020explainable, rudin2019stop}.
Notably, SI-MIL is generic enough to substitute any state-of-the-art MIL method in the MIL branch.
% , thus enabling to enhance MIL interpretability without compromising its predictive performance. 
In summary, our main contributions are:

\begin{itemize}
    % \item SI-MIL, the first interpretable-by-design MIL method for gigapixel WSIs, which provides improved patch-level interpretations compared to state-of-the-art MIL methods and de novo
    % feature-level interpretations grounded on PathExpert insights for a WSI. 

    \item SI-MIL, the first interpretable-by-design MIL method for gigapixel WSIs, which provides de novo
    feature-level interpretations grounded on pathological insights for a WSI.

    \item A novel co-learning strategy for SI-MIL to mitigate the model performance-interpretability trade-off associated with self-interpretable methods. We quantitatively establish the efficacy of our method for classification tasks on three cancer types.

    \item We demonstrate the utility of SI-MIL's local WSI-level and global cohort-level explanations thorough quantitative and qualitative benchmarking in terms of statistical analysis, a domain expert study, and desiderata of interpretability, \ie, \textit{fidelity}, \textit{user-friendliness} and \textit{faithfulness}.

    % \item A new dataset comprising nuclei maps, PathExpert features, attention maps derived from SI-MIL, and feature contribution score for $\approx2.2$K WSIs.

    \item We provide a comprehensive dataset for $\sim$2.2K WSIs, featuring nuclei maps, PathExpert features, and SI-MIL derived outputs, with the aim of streamlining the resource intensive preprocessing towards interpretability studies in computational pathology.
    %process of extracting handcrafted PathExpert features.
    
\end{itemize}

\section{Related work}
\label{sec:related work}

This section presents an overview of different forms of interpretability, primarily focusing on the domain of computational pathology.

\noindent \textbf{Post-hoc interpretability methods:}  These methods fall into two categories: patch-level and WSI-level. Patch-level techniques, like GradCAM and Layer-Wise Relevance Propagation (used in~\cite{hagele2020resolving, sadafi2023pixel}), highlight key pixels in model predictions. For deeper insight, studies like~\cite{jaume2020towards, jaume2021quantifying} use biological entity-based graphs for pathologist-friendly explanations. At the WSI-level, interpretability is primarily achieved through attention maps that identify salient regions in WSIs. Additionally, few methods, such as those by ~\cite{zheng2022graph, pati2023weakly}, use segmentation maps or gradient-based techniques to localize significant areas. However, these methods, as~\cite{rudin2019stop} notes, may suffer from a disconnect from the model's computations. In pathology, this is particularly evident when comparing the deep features used for MIL training and the handcrafted features used for subsequent analysis ~\cite{mckenzie2022interpretable, chen2022pan, fremond2023interpretable}, revealing a disparity in the features for training versus those for feature correlation.

% However, as highlighted in the seminal work~\cite{rudin2019stop}, post-hoc methods can suffer from impaired reasoning due to their independence from the model’s actual computations. For context in pathology, the post-hoc PathExpert feature correlation methods such as~\cite{mckenzie2022interpretable, chen2022pan, fremond2023interpretable}, use the deep features for MIL training and then PathExpert features for statistical analysis in identified salient region. Therefore there is clear evident disconnect between the features used for actual feature correlation and discovery (PathExpert) and the one used in training of MIL alone driving the selection of salient regions. 

% In the context of the above post-hoc methods~\cite{mckenzie2022interpretable, chen2022pan, fremond2023interpretable}, this independenc e is evident from the disconnect between the features used for MIL training, i.e., deep features, and those analyzed by statistical methods, i.e., PathExpert features. 

% Lime shapely for feature importance (https://www.nature.com/articles/s41598-020-62724-2)

% Regression Concept Vectors for Bidirectional Explanations in Histopathology
% Concept attribution: Explaining CNN decisions to physicians
% jain2018retinal

\noindent \textbf{Vision-Language methods:} Previous works~\cite{radford2021learning, lu2023towards, huang2023visual, li2023llava} have explored  interpretability using task reasoning through textual descriptions or vision-language similarity~\cite{aflalo2022vl, lu2023visual, zhang2017mdnet}. However these methods~\cite{lu2023visual, huang2023visual} either suffer from post-hoc approximation~\cite{rudin2019stop}, or are not scalable~\cite{zhang2017mdnet, li2023llava} for gigapixel images. Note that in pathology, most paired image-text data are only at the patch level~\cite{huang2023visual}. This makes it challenging to design WSI-level interpretable prediction models from only patch-level descriptions. Furthermore, at WSI-level, unlike natural images, the text descriptions in pathology reports are not holistic; i.e. these reports do not capture the complete landscape and primarily consist of global summaries of the pathologists' findings.

\noindent \textbf{Self-interpretability methods:} A family of models, grounded in concepts~\cite{koh2020concept, espinosa2022concept, oikarinen2023label, yuksekgonul2022post, yang2023language}, has become prominent for natural image interpretation. These models learn interpretable embeddings by mapping visual representations to a concept layer, and linearly aggregate these for prediction. Challenges such as information leakage and semantic inaccuracies are noted in ~\cite{margeloiu2021concept, mahinpei2021promises, barbiero2023interpretable}.
To address this, ~\cite{barbiero2023interpretable} uses concept embeddings ~\cite{espinosa2022concept} to learn syntactic rules and make predictions based on concept truths. While effective in interpretability, its validation is confined to Boolean logic tasks. Despite its emergence in the analysis of natural images, this field has yet to be explored in the context of gigapixel pathology. Building upon this, our proposed SI-MIL can handle complex WSI tasks while embedding interpretability directly into the MIL framework.
\vspace{-0.1cm}

\section{Method}
\label{sec:methods}

% \pp{I would move tings around here. First, lets explain the key contribution, i.e the interpretable MIL, then you can explain the feature extraction and MIL.}

\begin{figure*}[ht]
\centering

    \includegraphics[width=0.98\linewidth]{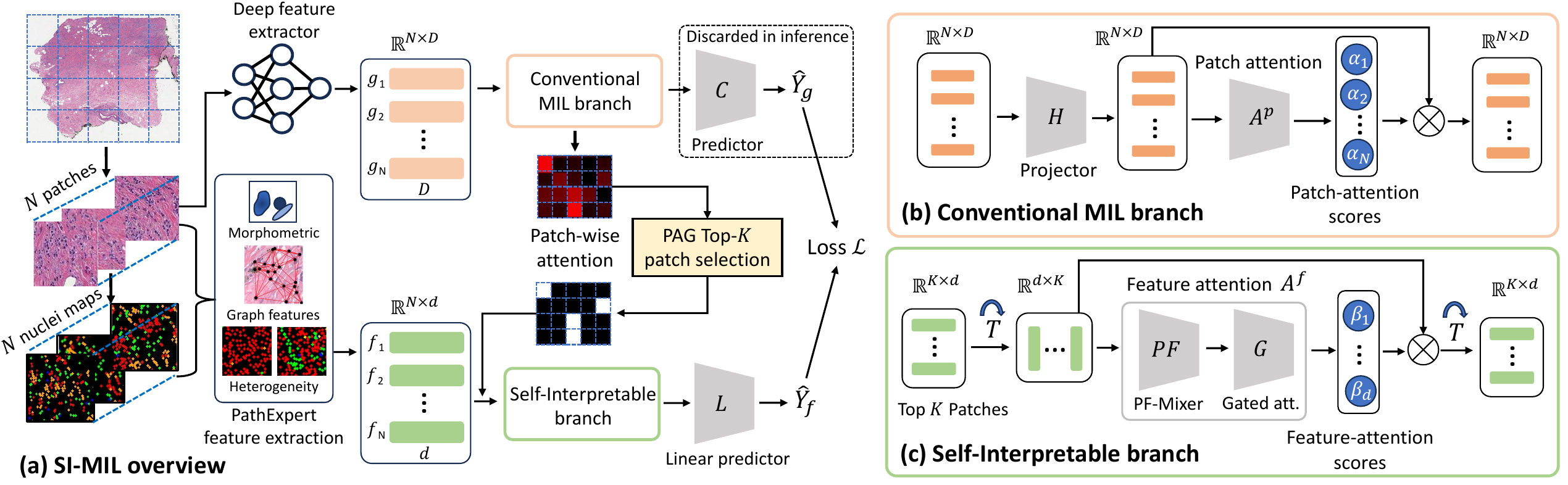}
    \caption{\textbf{Overview of SI-MIL}: Conventional MIL branch guides the \textit{Patch Attention-Guided Top-$K$} (PAG Top-K) patch selection module to select the PathExpert features of key regions from WSI, followed by linear scaling in the \textit{Self-Interpretable} branch, and linear prediction.}
    \label{fig:framework}
    \vspace{-0.5cm}
    
\end{figure*}

In this section, we present the details of our dual branch SI-MIL (overview in Fig.~\ref{fig:framework}), consisting of a conventional MIL branch and a \textit{Self-Interpretable} (SI) branch, for analyzing WSIs. We describe the conventional MIL in Sec.~\ref{Conventional-MIL}. Detailed description of the feature extraction pipelines, \ie, the process of extracting black-box deep features ($g$) and interpretable PathExpert features ($f$), is provided in Sec.~\ref{feature_extraction}. Finally, we present the complete SI-MIL framework in Sec.~\ref{SI-MIL}.

% Finally, we will present the setup for class-level feature discovery and slide-level interpretable prediction analysis unlocked using our proposed method.

%-------------------------------------------------------------------------
\subsection{Conventional MIL}
\label{Conventional-MIL}

% \pp{This looks a bit out of place. Maybe because of the subsection title. This title wold be something one would have in the related work - illustrating that yours is `unconventional' I think what you're trying discuss here is the upper part of the pipeline. If that's the goal, I would say up front that our SI-MIL pipeline consists of two main modules. Module 1 - SI branch. Module 2- A-MIL branch. So, first you discuss Module 1 which is the key contribution. And then Module 2, followed by feature extraction. Right before 3.1, you provide a brief description of the different modules before starting to expand on them.}

% Traditionally, MIL is used for WSI-level weakly-supervised  tasks~\cite{ilse2018attention,lu2021data,shao2021transmil}. Here, 
In conventional MIL, each WSI is decomposed into patches ($p_1$, $p_2$, \dots $p_N$), and their extracted features ($g_1$, $g_2$, \dots $g_N$), $ g_i \in \mathbb{R}^{D}$ are treated as a bag of instances.
% Towards the objective of devising an interpretable MIL,
In this work, we leverage an additive version of MIL~\cite{javed2022additive} in the conventional MIL, which imparts better spatial credit assignment to tissue regions in a WSI.
As illustrated in Fig.~\ref{fig:framework}\textcolor{red}{b}, conventional MIL consists of a projector $H(\cdot)$ operating on the input feature space, followed by a patch attention module $A^p(\cdot)$ to compute soft attention $\alpha$ over patches as follows: 
\vspace{-0.2cm}
\begin{equation}
\vspace{-0.2cm}
\tilde{g}_i = H(g_i); \hspace{0.5 cm}\alpha_i = A^p(\tilde{g}_i); \hspace{0.5 cm}i \in \{1,2,...N\}
\end{equation} 
\noindent where $A^p(\cdot)$ is a parameterized module with softmax activation. The attention-scaled feature embeddings are input to the predictor $C(\cdot)$ which estimates the marginal contribution of each patch to the slide-level task. Finally, these contributions are aggregated and activated with $\psi$ to infer slide-level prediction $\hat{Y}_g$ as:
\vspace{-0.2cm}
\begin{equation}
\vspace{-0.2cm}
\hat{Y}_g = \psi \Big( \sum_{i=1}^{N} C(\alpha_i \cdot \tilde{g}_i) \Big)
\end{equation}
\noindent The MIL performs slide-level prediction while computing the contributions through patch-level attentions. However, these attentions are too coarse for pathological interpretability as they do not explain the underlying patterns in pathologist-friendly terminologies.
%limiting the pathologists' ability to understand the underlying patterns contributing to the slide-level prediction. 

\subsection{WSI patch feature extraction}
\label{feature_extraction}
For each WSI, we extract patches ($p_1$, $p_2$, \dots $p_N$) and derive two sets of features for each patch $p_i$, defined as:

\begin{enumerate}
    \item \textbf{Deep features:} We pretrain a ViT~\cite{dosovitskiy2020image} through self-supervised learning on patches from the WSIs, and use the ViT as feature extractor to encode a patch $p_i$ into a deep feature vector $g_i \in \mathbb{R}^D$. 
    Note that any other pretrained or foundational model~\cite{wang2022transformer, chen2022scaling, vorontsov2023virchow, kapse2023attention, wang2023retccl, riasatian2020kimianet} can be used for patch encoding.
    
    \item \textbf{Hand-crafted PathExpert features:} We use HoVer-Net~\cite{graham2019hover}, pretrained on PanNuke~\cite{gamper2019pannuke} dataset, to segment and classify nuclei into 5 classes in each $p_i$. Then, pathologist-friendly features $f_i \in \mathbb{R}^d$ are extracted to quantify nuclei morphology and spatial distribution properties in $p_i$. These features can be grouped as:

% \begin{itemize}
    % \item 
    \underline{\textit{Morphometric properties}}, \ie, intensity, shape, and texture, are computed for all the nuclei in a patch, and are aggregated via statistical measures, \ie, mean, standard deviation, skewness and kurtosis for each nuclei class.
    
    % \item 
    \underline{\textit{Spatial distribution properties}} of different communities of nuclei types in a patch are quantified using graph analysis and heterogeneity. The former uses  nuclei centroids to construct cell-graph and then extracts social network analysis~\cite{zamanitajeddin2021cells}-based features for each nucleus, followed by statistical aggregation. These features capture properties such as degree of cohesiveness and nuclei clustering. The latter quantifies the spatial interaction of different nucleus class communities by using the nuclei centroids and class labels. Entropy and infiltration based descriptors are leveraged for this computation~\cite{martinelli2022athena}.     
    
    % For each cell, social network analysis based statistical features are computed to gauge its spatial configuration within a patch~\cite{zamanitajeddin2021cells}. These features are capable of capturing the spatial distribution properties of cells such as cohesive vs. non-cohesive, clustered vs. dispersed, etc. The derived features are then aggregated using statistical methods such as max, mean, std, skew, and kurtosis.
    
     % \item \textit{Spatial heterogeneity quantification}: Beyond the spatial configuration properties discussed above, this group of features utilizes not just the centroids and segmentation of cells, but also their classes, to quantify the spatial heterogeneity of different cell communities within a patch~\cite{martinelli2022athena}. Entropy-based descriptors quantify the uniformity versus randomness of cell communities at the global patch level. At the local level, we derive interaction and entropy scores for each cell using its neighbors in a k-NN graph. This approach captures the extent of spatial intermixing or clustering of cells from distinct classes. Clustered arrangements result in a prevalence of cells with lower entropy, as their neighbors are predominantly from the same class. In contrast, intermixed arrangements elevate entropy values, as neighboring cells are often from different classes.
    
% \end{itemize}

\end{enumerate}

\noindent The comprehensive list detailing the different groups of features, along with illustrative sample images are provided in supplementary. 
%These samples elucidate the significance of the features in terms of their PathExpert rationale.

%-------------------------------------------------------------------------
\subsection{Self-Interpretable MIL (SI-MIL)}
\label{SI-MIL}

As shown in Fig.~\ref{fig:framework}\textcolor{red}{a}, along with the aforementioned conventional MIL as a branch, SI-MIL consists of a \textit{Patch Attention-Guided Top-$K$} (PAG Top-$K$) module and a SI branch. The PAG Top-$K$ module aims for a differentiable selection of top $K$ patches identified in the MIL branch; thus enabling the co-learning with the SI branch. This branch operates on these top $K$ patches, by leveraging a feature attention module to linearly scale the corresponding relevant PathExpert features. The patch-wise PathExpert features and feature attention scores are subsequently aggregated by a linear predictor for slide-level task.
% The goal of this module is to contextualize the input features with multiple patches and feature mixing layers (\textit{PF-Mixer}), followed by an introduction to a gated attention module. 
% The core of SI-MIL lies the strategy of leveraging the robust discriminative power of deep features to guide the selection of high interest patches. 
% Self-Interpretable branch then performs linear scaling on the PathExpert features of these selected patches. The linearly scaled features are then fed to the linear predictor. 
SI-MIL denotes the dual branch co-learning framework that discriminates complex WSIs using a linear equation, advancing interpretability by introducing feature-level insights while maintaining high performance and complementing existing MILs. The details of the individual components are described in the following sections.

\textbf{PAG Top-$K$ patch selection module:} This module leverages the patch attention scores $\alpha$ from the conventional MIL branch to select the $K$ most salient patches in a WSI. As the naive Top-$K$ operation is non-differentiable, we use the differentiable \textit{perturbed Top-$K$} operation from~\cite{cordonnier2021differentiable, thandiackal2022differentiable}. This \textit{perturbed Top-$K$} operation is imperative to enable the co-learning of both the branches: conventional MIL and \textit{SI} branch. Following patch selection, only the PathExpert features of the salient patches are utilized in the subsequent steps. Therefore, the use of deep features in the MIL branch does not hinder the interpretability of SI-MIL; rather it guides the selection of informative patches, which is denoted as:
\vspace{-0.2cm}
\begin{equation}
\vspace{-0.2cm}
S_K = \text{TopK}(\alpha, K)
\end{equation}
where \( S_K \) denotes the indices of the selected top $K$ patches.

\textbf{Feature Attention module:} The $A^f(\cdot)$ module consists of a patch feature mixing network and gated attention network. Their synergistic integration forms a learnable feature selector without interfering with the interpretability of SI-MIL. First, the PathExpert feature matrix $M \in \mathbb{R}^{K \times d}$, corresponding to the \( S_K \) patches, is transposed and fed to a patch feature mixing network \textit{PF-Mixer}, ${PF}(\cdot)$. It contextualizes each value in \(M^T\) with the top $K$ patches and $d$ features. In practice, ${PF}(\cdot)$ is implemented via MLP layers~\cite{tolstikhin2021mlp}, with separate layers dedicated to mixing spatial patch information and per-patch feature information. 
%It consists of multiple \textbf{P}atch mixing and \textbf{F}eature mixing MLP layers~\cite{tolstikhin2021mlp} to contextualize each value in \(M^T\) with the top $K$ patches and $d$ features respectively. 
Subsequently, gated attention network $G(\cdot)$  processes each row of the matrix $M^T \in \mathbb{R}^{d \times K}$ independently to determine the attention score $\beta_j$ for each feature $d_j$, computed as:
\vspace{-0.2cm}
\begin{equation}
\vspace{-0.2cm}
\beta_j = G({PF}(M^T)); \hspace{0.5cm} j \in \{1,2,...d\}
\end{equation}
\noindent To enforce the model to be dependent on most salient features, we scale the feature attention scores $\beta$ as follows:  $\beta$ values  are first scaled using percentile  $Pr_\gamma$ (where $\gamma$ is the $\gamma^{th}$ precentile) and standard deviation (std), and then sigmoid activated with a hyper-parameter, temperature ($t$) as shown in Eqn.~\ref{b_scale}. 
This operation enforces the $\beta$ values above $Pr_\gamma$ towards 1 and remaining towards 0, thereby imposing sparsity in feature selection. Note that for brevity, we denote the scaled values of $\beta$ with same notation in Eqn.~\ref{b_scale}. 
\vspace{-0.2cm}
\begin{equation}
\vspace{-0.2cm}
\label{b_scale}
\beta_j = \frac{\beta_j - Pr_\gamma(\beta)}{\text{std}(\beta)}; \hspace{0.5 cm} \beta_j = \frac{1}{1 + e^{-\beta_j \times t }}
\end{equation}
\noindent These feature attention values are used to linearly scale the PathExpert feature matrix $M$ such that the salient features are emphasized while attenuating others:
\vspace{-0.2cm}
\begin{equation}
\vspace{-0.2cm}
\label{eq:linearscale}
M^{'}_{ij} = \beta_j \times M_{ij}; \hspace{0.5cm} i \in \{1,2,...K\}; \hspace{0.1cm} j \in \{1,2,...d\}
\end{equation}
\noindent Note that even though $A^f(\cdot)$ includes non-linear operations to compute $\beta$, the original feature space $M \in \mathbb{R}^{K \times d}$ is just linearly scaled with $\beta$. $A^f(\cdot)$ paves the way for linear prediction in the next stage, while preserving interpretability.
%Please note that this module outputs the linearly scaled pathologica feature space of the selected patches $M$, thereby preserving the interpretability of the feature space, as described in eq.~\ref{eq:linearscale}.
%These operations paves the way for linear prediction in the next stage.

\textbf{Linear Predictor and Aggregation:} Following the attention scaling of the PathExpert features corresponding to the \( S_K \) patches, the features are fed to a linear predictor $L(\cdot)$ characterized by weights \(w(\cdot)\) and bias \(b\) as:
% M^{w\beta}_{ij} = w_j \times M^\beta_{ij}; \hspace{0.5 cm}
\vspace{-0.2cm}
\begin{equation}
\vspace{-0.2cm}
\label{equation:m_wb}
M^{''}_{i} = \sum_{j=1}^{d}w_j M^{'}_{ij} + b; \hspace{0.5cm} i \in \{1,2,...K\}
\end{equation}
\noindent Finally, for slide-level prediction, the contributions $M^{''}_{i}$ of the selected patches undergo an aggregation and an activation $\psi$ as:
\vspace{-0.2cm}
\begin{equation}
\vspace{-0.1cm}
\label{equation:yf_pred}
\hat{Y}_f =  \psi \Big( \sum_{i=1}^{K} M^{''}_{i}\Big)
\end{equation}
\noindent It can be observed that the WSI-level prediction in the SI branch can be decomposed into a \textit{linear combination} of feature attention scores $\beta$, classifier weights \(w(\cdot)\), and the PathExpert feature matrix $M \in \mathbb{R}^{K \times d}$ of the top $K$ patches (\( S_K \)), given as:
\vspace{-0.2cm}
\begin{equation}
\vspace{-0.2cm}
\label{equation:linear}
\hat{Y}_f =  \psi \Big( \sum_{i=1}^{K} \sum_{j=1}^{d} w_j  \beta_j  M_{ij} + b\Big)
\end{equation}

\textbf{Optimization:}
Given the true label $Y$ for a WSI, the predictions $\hat{Y}_g$ from the MIL branch and $\hat{Y}_f$ from the SI branch, SI-MIL is optimized using slide-level cross entropy losses $\mathcal{L}_{CE}$ for both the predictions with respect to the true label. This joint optimization tames the patch attention module to select the patches collaboratively in the deep feature and PathExpert feature space. To enhance the performance of the SI branch, a knowledge distillation loss $\mathcal{L}_{KD}$ is optimized based on the mean squared error between $\hat{Y}_f$ and $\hat{Y}_g$. $\mathcal{L}_{KD}$ enforces alignment in performance between the two branches. The overall loss is computed as:
\begin{equation}
\mathcal{L} = \mathcal{L}_{CE}(Y, \hat{Y}_g) +  \mathcal{L}_{CE}(Y, \hat{Y}_f) + \lambda 
\mathcal{L}_{KD}(\hat{Y}_g, \hat{Y}_f)
\end{equation}
\noindent where $\lambda$ is used as a weight to align the scale of $\mathcal{L}_{KD}$ with the $\mathcal{L}_{CE}$ losses of deep feature and self-interpretable branch. Note that during inference, prediction from any branch can be used. However to enforce interpretability, the WSI-level prediction  is obtained from the SI branch, \ie, $\hat{Y}_f$ is considered for slide-level prediction and the non-interpretable branch's output $\hat{Y}_g$ is discarded.

\vspace{-0.1cm}

% the MIL branch only provides the patch-level attentions, whereas  

%-------------------------------------------------------------------------

%-------------------------------------------------------------------------

\section{Experiments: Prediction Performance}
\label{sec:results}
\vspace{-0.1cm}

Here, we first describe the datasets and implementation details, common to both performance and interpretability assessment. Then, we benchmark SI-MIL on multiple WSI classification tasks. We conclude with ablation studies and showcasing adaptability of SI-MIL to various MIL models. 

% Note that in our model, both branches can be used to make prediction. Here in this section we intentionally evaluate prediction power of self-interpretable branch. The prediction power of MIL branch is atleast as good as self-interpretable branch. 

\subsection{Datasets and Implementation details}
\label{dataset_and_implementations}

\hspace{0.4cm} \textbf{Datasets:} We evaluate SI-MIL on three WSI datasets: TCGA-BRCA~\cite{tcga_brca}, TCGA-Lung~\cite{tcga_luad, tcga_lusc}, and TCGA-CRC~\cite{tcga_coad}. \textbf{TCGA-BRCA} contains 910 diagnostic slides of two breast cancer subtypes: invasive ductal carcinoma (IDC) and invasive lobular carcinoma (ILC). \textbf{TCGA-Lung} contains 936 slides of two non-small cell lung cancer subtypes: lung adenocarcinoma (LUAD) and lung squamous cell carcinoma (LUSC). \textbf{TCGA-CRC}: includes 320 slides of colorectal cancer with low- or high-mutation density for hypermutation. Additional details about train-test splits are provided in the Supp. Sec.~\ref{additional_dataset_section}.

\textbf{Patch and feature extraction:} Patches of size $224 \times 224$ at 5$\times$ magnification and corresponding $1792 \times 1792$ at 40$\times$ are extracted for each dataset. For deep features extraction, we pretrain ViT-S~\cite{dosovitskiy2020image} with DINO~\cite{caron2021emerging} on the 5$\times$ patches from the training splits of individual datasets mentioned above. PathExpert features are extracted on corresponding patches at $40\times$. 

\textbf{MIL setting:} Additive ABMIL~\cite{javed2022additive} is adopted as the conventional MIL in this study. $A^p(\cdot)$ and $A^f(\cdot)$ are deep neural network based gated attention modules adopted from~\cite{lu2021data}. For all the MIL experiments, the batch size is set to 1 to handle WSIs of variable bag sizes. For robustness, 5-fold cross-validation is performed on the train split and the mean performance on the held-out test split is reported. By default, $\#$PF-Mixer layers $=$4, $\lambda = 20$, $K = 20$, $\gamma=0.75$, and $t=3$. More implementation details are provided in the Supp. Sec.~\ref{additional_implementation_section}.
Note that SI-MIL is evaluated with only DINO ViT-S features. Experimentation with other deep features is left for future exploration.

\vspace{-0.1cm}
\subsection{Slide-level classification performance}
% \cc{Fidelity is confusing (I thought it is connected to interpretability). I'd rather use accuracy. In fact if you really want to talk about fidelity, accuracy is not enough. You might have to compare the learnt attention map, patch-wise prediction of the PathExpert model, etc, with those of a pure MIL model. This means the explanation is faithful to the original MIL model.}
In this section, we benchmark the WSI classification performance of SI-MIL in terms of accuracy and area under the curve (AUC), which are the commonly employed metrics to quantify the \textit{fidelity} of interpretability algorithms~\cite{graziani2023global}. 
Table~\ref{table:results} presents the classification performance of SI-MIL and the competing baselines.
In absence of WSI-level self-interpretable methods, we construct interpretable baselines by perturbing SI-MIL under various settings. The baselines can be grouped in terms of the types of employed features as follows:
 
\textbf{Baselines using deep features:} \label{deep_baselines} These baselines denote training Additive ABMIL with features from different pretrained deep feature extractors, \ie, ImageNet~\cite{deng2009imagenet} supervised ViT-S (IN ViT-S), RetCCL~\cite{wang2023retccl}, CTransPath~\cite{wang2022transformer}, and our pretrained DINO ViT-S.  Although these baselines can render patch-level contributions in terms of attention maps, one cannot entirely deduce the reasoning behind these patch attentions, and cannot obtain feature-level understanding due to their inherently non-interpretable characteristics.

\textbf{Baselines using PathExpert features:} These baselines denote training Additive ABMIL with PathExpert features. To induce interpretability, we train MIL with PathExpert features, referred as PathFeat. However, this framework is non-interpretable as the projector $H(\cdot)$ maps the PathExpert features into a non-interpretable deep feature space. Therefore, we include a true interpretable baseline by training the MIL without  $H(\cdot)$.

\textbf{2-stage training using PathExpert features:}
Here, we first train the Additive ABMIL and extract top-$K$ attended patches for each WSI. Then, a self-interpretable linear classifier using the PathExpert features from the patches is trained. Specifically, we train the SI branch independent of the MIL branch, \ie, without PAG Top-$K$. This is analogous to the post-hoc analytical methods in~\cite{fremond2023interpretable, chen2022pan}. 

%\textbf{Ours:} Finally, we evaluate our proposed SI-MIL on the three datasets and compare it to baselines defined above. 
% Finally, we investigate two training  strategies for our proposed SI-MIL: (i) a 2-stage training approach, and (ii) joint training of the \textit{conventional MIL} and SI branches. 
% In the 2-stage training approach, we first train the \textit{conventional MIL branch}, followed by the extraction of the top-K attended patches for each WSI. Thus, there is no gradient flow from the \textit{Self-Interpretable branch} to the \textit{conventional MIL branch}.
% Subsequently, the \textit{Self-Interpretable branch} is trained using the PathExpert features of these selected patches.
%In the SI-MIL, both \textit{conventional MIL} and SI branches are optimized simultaneously. In inference, the performance is assessed using the output \(\hat{Y}_f\) from the \textit{Self-Interpretable branch}. 
% Note that the default configuration for SI-MIL is the joint training strategy, while the 2-stage training approach is investigated specifically to emphasize the importance of joint training, as discussed next.

 % \multirow{2}{*}{SSL framework}
\begin{table}[!t]
\caption{Results indicate the mean of 5-fold cross-validation on test set. All methods are trained with Additive ABMIL as base MIL. Int. denotes self-interpretability of a method.}
\vspace{-0.4cm}
\label{table:results}
% \tiny
\begin{center}
\resizebox{\columnwidth}{!}{
\begin{tabular}{cccccccc}
\toprule
        &      & \multicolumn{2}{c}{\textbf{Lung}} & \multicolumn{2}{c}{\textbf{BRCA}} & \multicolumn{2}{c}{\textbf{CRC}}        \\ 
          & Int.    & \multicolumn{1}{c}{Acc.}       & AUC    & \multicolumn{1}{c}{Acc.}       & AUC   & \multicolumn{1}{c}{Acc.}       & AUC     \\  
\midrule

IN ViT-S    & \xmark         &    0.859  &    0.919   &    0.929  &    0.967   &    0.891  &    0.898   \\ 

   RetCCL          & \xmark     &      0.860 &    0.935   &    0.929  &    \textbf{0.976}   &    0.889  &    0.891    \\ 

  CTransPath          & \xmark     &      \textbf{0.904} &    \textbf{0.967}   &    0.920  &    0.974   &    \textbf{0.906}  &    \textbf{0.897}    \\ 

  DINO ViT-S     & \xmark   &      0.896  &    0.957   &    \textbf{0.937} &    0.974   &    0.904 & \textbf{0.897}   \\ 

\midrule

PathFeat       & \xmark    &      0.830  &    0.888   &    0.885  &    0.950   &    \textbf{0.886}  &    0.818     \\ 

 PathFeat w/o $H(\cdot)$        & \cmark    &      0.767   &    0.837   &    0.889  &    0.914   &    0.853   &    0.720   \\ 

 2-stage training & \cmark  &      0.865  &    0.932   &    0.908  &    0.924   &    0.876    &    0.862   \\

   SI-MIL (ours)   & \cmark   &    \textbf{0.884}   &    \textbf{0.941}   &    \textbf{0.944}  &    \textbf{0.968}   &    0.884    &    \textbf{0.910}    \\ 

% \midrule
\midrule

 \multicolumn{8}{c}{\textbf{Ablation study of SI-MIL components}} \\
\midrule

  % SI-MIL   & \cmark   &    \textbf{0.884}   &    \textbf{0.941}   &    \textbf{0.944}  &    \textbf{0.968}   &    \textbf{0.884}    &    \textbf{0.910}    \\ 

 w/o  PAG Top-$K$  & \cmark  &     0.859  &    0.936   &    0.915  &    0.922   &    0.876    &    0.869    \\ 

% w/o  $A^f(\cdot)$     & \cmark   & 0.853  &    0.935   &    0.939  &    \textbf{0.981}   &    0.871  &    0.857    \\ 

%   w/o  ${Mix}(\cdot)$     & \cmark   & 0.838  &    0.915   &    0.925  &    0.953   &    0.866  &    0.863    \\ 

  w/o  KD   & \cmark   &    0.853   &    0.915   &    0.932  &    0.951   &    0.878  &    0.830     \\ 

 w/o PAG Top-$K$ \& KD & \cmark  &      0.857  &    0.924   &    0.915  &    0.899   &    0.879    &    0.858   \\

\bottomrule

\end{tabular}
}
\end{center}
\vspace{-0.8cm}
\end{table}

\textbf{Results:} As observed in Table~\ref{table:results}, conventional MIL using PathExpert features, and particularly the one without projector, performs considerably worse than the methods using deep features. This accuracy-interpretability trade-off often undermines the benefits of using interpretable frameworks/features. SI-MIL aims to close this performance gap by utilizing deep feature-based guidance. We find that SI-MIL, despite imposing a linear constraint (Eq.~\ref{equation:linear}) on predictions, elevates the performance of PathExpert features to be on par with deep feature-based baselines. 

Note that the results for RetCCL and CTransPath are potentially inflated as the feature extractors were pretrained on the entire TCGA cohort, including test splits used in our study. Thus, the DINO ViT-S and IN ViT-S baselines, unaffected by the test split, provide a more accurate comparison.

%This underscores the feasibility of predicting complex WSIs using a linear combination of PathExpert features from the most discriminative regions without performance compromise. 
% We also observe that the 2-stage training strategy under performs compared to the joint training strategy  in SI-MIL. %This indicates the necessity of optimizing the patch attention module $A^p(\cdot)$ by taking into account the influence of  both deep features and PathExpert features. 

\begin{figure*}[ht!]
\centering
    \includegraphics[width=0.96\linewidth]{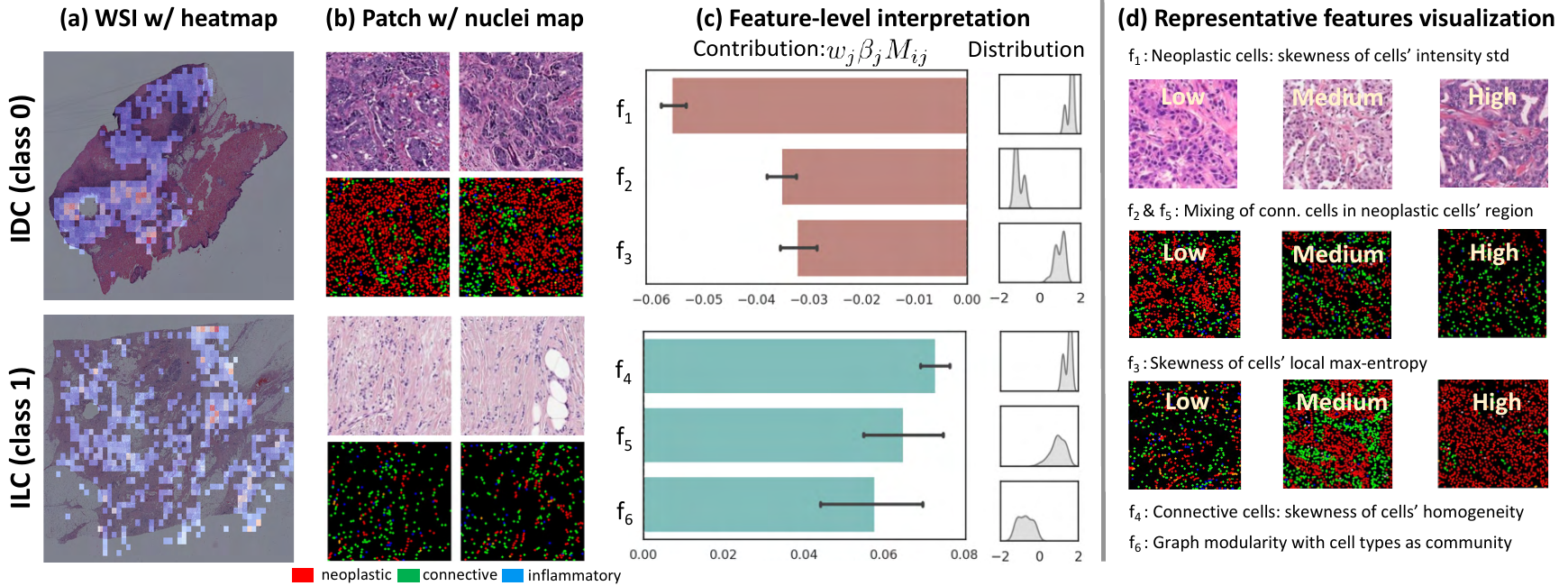}
    \caption{\textbf{Qualitative Patch-Feature importance report:} In (a) and (b), we present WSIs with overlaid attention heatmaps and the top two patches, along with their nuclei maps. In (c), we demonstrate the mean contribution magnitude of select representative features across the top $K$ patches employed in the Self-Interpretable branch. Additionally, we display a feature density plot that quantifies the distribution of features within the $K$ patches. For brevity, we omit the y-axis. Given that these features are normalized, a curve leaning towards the right indicates higher/positive values, while one towards the left signifies lower/negative values, depending on the feature. Finally, in (d), we illustrate, the description and visualization of representative features in (c) with varying value.}
    \label{fig:patient_interpretable}
    \vspace{-0.4cm}
\end{figure*}

\textbf{Ablation studies:} In Table~\ref{table:results}, we mainly showcase the significance of the meticulously designed components of SI-MIL. We show the implication of the PAG Top-$K$ module by omitting the perturbed Top-$K$ selection and blocking gradient flow from the SI branch to the MIL branch. In Table~\ref{table:results}, we observe that a non-differentiable approach degrades the performance. This indicates that the most discriminative region identified by the MIL is potentially less effective in the PathExpert feature space, thus highlighting the need to find regions discriminative in both feature spaces for enhancing the predictive power of SI branch. It can also be observed that in both settings, with or without perturbed Top-$K$, knowledge distillation is instrumental in enhancing the performance. $\mathcal{L}_{KD}$ acts as a regularizer for the SI branch, pushing it to stay as close as to the high-performing MIL branch.
Additional ablations demonstrating the effect of varying \( K \) in the PAG Top-$K$ module, the number of PF-Mixer layers, and the percentile and temperature for scaling \( \beta \) are presented in the Supp. Sec.~\ref{additional_ablation}.

\textbf{Adaptability of SI-MIL to other MILs:}
% Here we address the adaptation of the \textit{Self-Interpretable branch} in SI-MIL with the Additive version of CLAM, PromptMIL, and TransMIL, as opposed to the default Additive ABMIL used in this study.
On the TCGA-BRCA dataset, we evaluate the generalizability of SI-MIL by adapting to state-of-the-art MIL frameworks, \ie, ABMIL~\cite{ilse2018attention}, CLAM~\cite{lu2021data}, and TransMIL~\cite{shao2021transmil}, in the MIL branch. Results in Table~\ref{table:ablation} establish that our SI-MIL extensions remain competitive with the corresponding MIL methods using standalone DINO ViT-S features.

\begin{table}[!t]
\caption{Mean of 5-fold cross-validation for adapting SI-MIL with other MIL frameworks (additive versions~\cite{javed2022additive}) on TCGA-BRCA.}
\vspace{-0.4cm}
\label{table:ablation}
\small
\begin{center}
\begin{tabular}{ccccc}
\toprule
        &   \multicolumn{2}{c}{\textbf{DINO ViT-S}} & \multicolumn{2}{c}{\textbf{SI-MIL}}    \\ 
        
          MIL    & Acc.       & AUC    & Acc.       & AUC      \\  
 
\midrule
 ABMIL  & 0.937  &    0.974  &      0.944  &    0.968 \\
CLAM & 0.937  &    0.972 &      0.925  &    0.957 \\
TransMIL & 0.934  &    0.936 &      0.929  &    0.933 \\

\bottomrule

\end{tabular}
\end{center}
\vspace{-0.7cm}
\end{table}

% \vspace{-0.1cm}

\section{Experiments and Results: Interpretability} 
\label{sec_intepretability}
% \vspace{-0.1cm}

In this section, we evaluate our SI-MIL model across various statistical criteria, \ie, univariate and multivariate class-separability, and desiderata of interpretability~\cite{graziani2023global}, \ie, \textit{user-friendliness} and \textit{faithfulness}, focusing on both local slide-level and global cohort-level interpretations. The \textit{user-friendliness} metric evaluates how easily end-users, \ie pathologists, can understand and trust the model predictions, and the \textit{faithfulness} metric gauges the extent to which model's explanations align with the expert's reasoning. The paper includes detailed analyses on TCGA-BRCA test WSIs. Further analyses on TCGA-Lung and TCGA-CRC are presented in the Supp. Sec.~\ref{additonal_interpretability}.
%To this end, the interpretability outcomes of SI-MIL, on TCGA-BRCA test WSIs, are thoroughly assessed by a domain-expert study conducted with a pathologist. 
%
%We perform the analysis using TCGA-BRCA test WSIs that are correctly predicted by both SI-MIL and conventional Additive ABMIL trained with DINO ViT-S features.
%\textbf{Implementation details:} For this study we focus on interpretability analysis for TCGA-BRCA. The slides from test set which are correctly predicted by our SI-MIL and conventional MIL trained with DINO ViT-S are selected.

\subsection{Local Interpretation: Slide-level}
\vspace{-0.1cm}

SI-MIL can explain model predictions at WSI-level without relying on post-hoc methods~\cite{arrieta2020explainable}. Contrary to existing  MIL~\cite{lu2021data, shao2021transmil, javed2022additive}, SI-MIL can produce both patch- and feature-level explanations, due to the linear mapping between the PathExpert features and output predictions. 
Fig.~\ref{fig:patient_interpretable} presents aggregated patch-feature importance reports generated by SI-MIL for two TCGA-BRCA WSIs, elucidating the rationale behind the predictions. 
Below, we explain the setup for generating such reports and then quantify their quality in terms of \textit{user-friendliness} and \textit{faithfulness}.

\noindent \textbf{WSI-level patch-feature importance report setup:} Input WSIs with overlaid patch-attention saliency maps, generated by the MIL branch are shown in  Fig.~\ref{fig:patient_interpretable}\textcolor{red}{a}. Up next, Fig.~\ref{fig:patient_interpretable}\textcolor{red}{b} shows the informative top $K$ patches and their nuclei predictions ($K=$2 for simplicity).
% highlighting those with the most contribution (as referenced in $M^{w\beta}_{i}$ in eq.~\ref{equation:m_wb}) to the corresponding classes
The nuclear map identifies the nuclei types and highlights their spatial organization in the tissue.
Next, the feature contributions across the top $K$ patches are detailed in Fig.~\ref{fig:patient_interpretable}\textcolor{red}{c}. 
Recall that in Eqn.~\ref{equation:linear}, $w_j\beta_j M_{ij}$ denotes the contribution of the $i$-th patch and its $j$-th feature, where $\sum_{i=1}^{K} w_j\beta_j M_{ij}$ infers the aggregated contribution of the $j$-th feature towards WSI prediction. We present the mean contributions and  95\% confidence intervals across $K$ patches, shown only for the three most contributing features for simplicity.
The negative and positive contributions are indicative of class 0 (IDC) and class 1 (ILC), respectively, as activated by sigmoid in Eq.~\ref{equation:yf_pred}. 
The feature distribution shows the range of the corresponding normalized features across $K$ patches. The distribution inclining towards left or right indicates low/negative values or high/positive values of the feature, respectively.
Looking at the distribution and contribution together, if we have a positively inclining distribution and negative contribution, then it means increasing the feature pushes the prediction towards class 0.
Fig.~\ref{fig:patient_interpretable}\textcolor{red}{d} illustrates a few features identified in Fig.~\ref{fig:patient_interpretable}\textcolor{red}{c}, with the representative patches having low and high value of corresponding features. 

\noindent \textbf{User-friendliness:} 
We qualitatively evaluate the utility of the patch-feature reports by an expert pathologist. 
First, we presented top $K$ ($=20$) patches and corresponding nuclear maps for the WSIs in Fig.~\ref{fig:patient_interpretable}\textcolor{red}{a} to the expert. The selected IDC and ILC patches demonstrated good agreement with class-specific prior knowledge. The IDC patches contained coherent cancer cells forming malignant glands, nests, or sheets with commentary about nuclear size, shape, color, and chromatin texture; and the ILC patches showed infiltrating small round cells in single file configurations. 
Afterwards, the top 10 contributing PathExpert features and associated feature distributions, as identified by SI-MIL in Fig.~\ref{fig:patient_interpretable}\textcolor{red}{c}, were evaluated by the expert to assess their correlation with domain knowledge in classifying IDC and ILC. 90\% and 80\% of these features for IDC and ILC WSIs in Fig.~\ref{fig:patient_interpretable}\textcolor{red}{a} were found relevant, respectively, due to their strong association with cell cohesiveness, nuclear hyperchromaticity, and morphology of cancerous nuclei properties.
These analyses helped the pathologist in reasoning with the model's rationale and developing trust in model's predictions. Interestingly, the pathologist commented on the utility of such feature-level relevance report in downstream correlations with genomic and laboratory data.

\begin{figure*}[ht!]
\centering
    \includegraphics[width=0.96\linewidth]{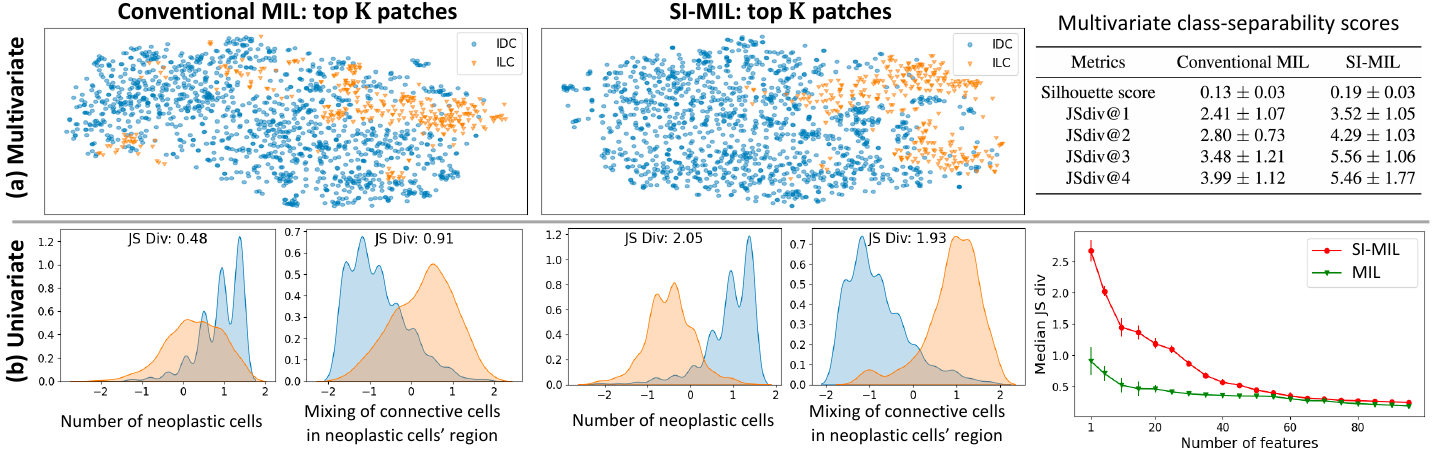}
    \vspace{-0.3cm}
    \caption{\textbf{Cohort-level Interpretation}: Separability of top $K$ patches of WSIs across classes in the PathExpert feature space. Multivariate and Univariate analyses depict that the top $K$ patches selected by SI-MIL and their PathExpert features are more separable.}
    \vspace{-0.4cm}
    \label{fig:cohort_tsne}
\end{figure*}

\iffalse
\begin{table}[ht]
\begin{center}
\resizebox{\columnwidth}{!}
{
    \begin{tabular}{ccc}
    \toprule
    Metrics & Conventional MIL  & SI-MIL     \\  
    \midrule
       Silhouette score  &    0.13 $\pm$ 0.03  &    0.19 $\pm$ 0.03        \\ 
       JSdiv@1           &    2.41 $\pm$ 1.07  &    3.52 $\pm$ 1.05        \\ 
       JSdiv@2           &    2.80 $\pm$ 0.73  &    4.29 $\pm$ 1.03        \\ 
       JSdiv@3           &    3.48 $\pm$ 1.21  &    5.56 $\pm$ 1.06        \\ 
       JSdiv@4           &    3.99 $\pm$ 1.12  &    5.46 $\pm$ 1.77        \\ 
    \bottomrule
    \end{tabular}
}
\end{center}
\end{table}
\fi

\noindent \textbf{Faithfulness:} We evaluated the faithfulness of our reports by quantifying the alignment of the top identified PathExpert features with pathologist's assessments. 
The evaluation involved the pathologist assigning relevance scores to the top features. 
Specifically, we selected 10 WSIs each from IDC and ILC, and generated patch-feature reports including top 10 contributing PathExpert features. Then, the reports were analyzed and the features were categorized into high-, moderate-, or non-relevant categories by the expert. 
The mean and standard deviation of the number of features in each category are reported separately in Table~\ref{tab:Pathologist_evaluation}. Also, an aggregated percentage of the number of features in each category is reported.
The analysis shows that the majority of the identified features are either highly or moderately relevant towards correct classification and interpretability. 
Among the non-relevant features, a few are interesting to be analyzed on larger cohorts to potentially discover new diagnostic biomarkers. The selection of some of the non-relevant features may also be due to certain misclassifications by HoVer-Net. This is left for future exploration.
% The current limitations of our method is over-reliance on HoVer-Net's prediction; thus incorrect cell classification particularly can cause SI-MIL to select more non-relevant features in some WSIs.

% , and the remaining can be considered incorrect due to nuclei misclassifications by HoVer-Net.

\begin{table}[ht]
\caption{Pathologist evaluation at slide-level for top contributing features' relevancy for IDC and ILC classes in TCGA-BRCA. Agg. denotes aggregated percentages of features belonging to three relevancy groups.}
\vspace{-0.5cm}
\label{tab:Pathologist_evaluation}

\begin{center}
\resizebox{\columnwidth}{!}{

\begin{tabular}{cccc}
\toprule
     &  Highly Relevant & Moderately Relevant  & Non Relevant     \\  
\midrule

   IDC & 5.40 $\pm$ 1.43  &    2.10 $\pm$ 0.94  &    2.50 $\pm$ 1.28           \\ 
   ILC & 3.25 $\pm$ 0.97  &    3.75 $\pm$ 0.83 &    3.00 $\pm$ 1.12          \\ 
\midrule

   Agg. & 44.5\%  &    28.3\%  &    27.2\%           \\ 

\bottomrule
\end{tabular}
}
\end{center}
\vspace{-0.6cm}
\end{table}

%\textcolor{red}{To Do: Asks Raj for a paragraph about these findings. RG: already explained in the paragraph, right? Saarthak mentioned expanding feature details in a supplement.

%RG: it's been edited, but I think that the sentence can be moved here, "The top features were associated with cell cohesiveness (e.g. spatial networks connectivity between cancer and neighboring stromal cells) and the color (e.g. hyperchromatic purple nuclei), size, shape, and texture of cancer nuclei.}

\subsection{Global Interpretation: Cohort-level}
\vspace{-0.1cm}

In this section, we holistically analyze how SI-MIL interprets at a global cohort-level and benefits over conventional MIL. Specifically, we perform univariate and multivariate statistical analysis to measure class-separability in the PathExpert feature space, inline with ~\cite{fremond2023interpretable, chen2022pan, mckenzie2022interpretable}.

\noindent \textbf{Univariate and Multivariate class-separability:} 
Through global cohort-level analysis, we demonstrate that SI-MIL, which includes the co-learning of MIL and SI branches, optimizes the selection of more informative patches than conventional MIL. 
During inference for both the models, we separately collect the top $K$ attended patches across WSIs corresponding to the two classes in TCGA-BRCA. Subsequently, we use the pre-extracted $d$ PathExpert features for the selected patches, as described in Sec.~\ref{feature_extraction}. Formally, given $N_1$ and $N_2$ number of WSIs in the two classes, we construct PathExpert feature matrices $F_1 \in \mathbb{R}^{(N_1 \times K) \times d}$ and $F_2 \in \mathbb{R}^{(N_2 \times K) \times d}$ for both the models.

\textbf{Multivariate} analysis employs t-SNE~\cite{van2008visualizing} to project $F_1$ and $F_2$ into a 2D embedding space, as shown in Fig.~\ref{fig:cohort_tsne}\textcolor{red}{a}. Afterwards, we measure the class-separability in terms of two metrics: (1) \textit{JSdiv@i}, which entails fitting a 2D Gaussian mixture model with $i$ components to each class and calculates the Jansen-Shannon (JS) divergence between the two distributions; and (2) \textit{Silhouette score}~\cite{rousseeuw1987silhouettes}, an unsupervised metric to evaluate the quality of class-wise created clusters. 
Both the measures are distance-based metrics that aim to highlight how separable the patches from the two classes are, in the projected embedding space. 
To account for modeling variability, we report the mean and standard deviation of the metrics across 5-fold cross-validation, as presented in Sec.~\ref{dataset_and_implementations}.
It can be observed in the table in Fig.~\ref{fig:cohort_tsne}\textcolor{red}{a} that SI-MIL consistently provides higher class-separability scores than conventional MIL method. This can be attributed to the co-learning technique in SI-MIL, which results in selecting more informative patches for individual classes that are better separable in the PathExpert feature space.

\textbf{Univariate} analysis examines the class-separability of patches for individual PathExpert features. For a given feature, \ie, a column in $F_1$ and $F_2$, we create class-wise density distributions and measure the JS divergence. For visual simplicity, we show the univariate analysis for the two PathExpert features for both SI-MIL and the conventional MIL in Fig.~\ref{fig:cohort_tsne}\textcolor{red}{b}. We can observe that the class-wise density distributions in SI-MIL are significantly better separated than the MIL. This further supports our argument of better patch selection in SI-MIL from multivariate analysis. 
For an aggregated univariate analysis, we rank the features by the decreasing order of JS divergence, and plot the median JS divergence against the increasing number of features. Similar to multivariate analysis, we state the mean and std of the medians across 5-fold cross-validation (Fig.~\ref{fig:cohort_tsne}\textcolor{red}{b}). 
We can observe that SI-MIL provides significantly better median class-separability for a good number of features, which strongly supports the enhanced quality of selected patches while preserving pathological understanding.

\subsection{Dataset contribution}
\vspace{-0.1cm}

We contribute a comprehensive dataset aimed at enhancing interpretability and reproducibility in MIL research. It comprises of nuclei maps, PathExpert features, and SI-MIL-generated patch-feature importance reports for 2.2K WSIs. 
% The creation process involved analyzing gigapixel WSIs at 40x magnification, utilizing HoVer-Net~\cite{graham2019hover} for cell segmentation and classification, followed by extraction of PathExpert features and feature importance scores detailed in Sec.~\ref{feature_extraction} and Sec.~\ref{sec_intepretability}. 
WSI processing and feature extraction involved significant computing resources (details in Supp. Sec.~\ref{data_contribution}). The complete list of PathExpert features, including cell shape and texture properties, spatial configurations, and interactions among different cell types, is detailed in the supplementary material.
% each WSI required about two hours, divided between GPU-based cell map prediction and CPU-based PathExpert feature extraction. Employing three RTX 8000 GPUs and a 40-core CPU with 500GB RAM, the total processing time amounted to roughly 4400 hours, or approximately 60 days.
% This list can be expanded to incorporate additional task-driven features extracted from the provided cell maps. 
We provide the key elements to enable researchers to further expand on the already comprehensive set. 

% and acknowledge the absence of tissue segmentation models and more refined methods for detecting cell classes.
\vspace{-0.2cm}

\section{Conclusion}
\label{sec:conclusion}
\vspace{-0.1cm}
We present Self-Interpretable MIL, which not only augments model interpretability by identifying salient regions and providing feature-level contributions within these regions but also achieves high performance on gigapixel WSI tasks. SI-MIL bridges the gap between AI-driven analysis and pathologist-friendly reasoning, a first of its kind in histopathology. From an evolutionary perspective, different cancers may share fundamentally similar characteristics; the PathExpert features in SI-MIL can capture these properties, possibly lending itself well to rare/unseen cancers. Future work will also involve integration of LLM-driven pathological concepts in model training. 
\vspace{-0.1cm}

\section{Acknowledgments}
\label{sec:acknowledgments}
\vspace{-0.1cm}
Reported research was supported by NIH 1R21CA258493-01A1, NSF IIS-2123920, IIS-2212046, and the Stony Brook Profund 2022 grant.

% We believe that comp

% Our findings indicate a promising direction for the development of reliable and interpretable AI tools in pathology. 

% Grounded by biology can potentially imporve model's generalization? hint for this. Will try this experiment for supplementary, if good then we can put it. SO need to subtly list it here

% Briefly talk about how LLMs with knowledge of geometry and PathExpert concepts can use our features to directly do the prediction. All language models followin feature extraction? 
% \input{sec/2_formatting}
% \input{sec/3_finalcopy}
{
    \small
    \bibliographystyle{ieeenat_fullname}
    \bibliography{main}
}

\clearpage
\setcounter{page}{1}
\maketitlesupplementary

In this supplementary material,  details are provided on the following: 
\begin{itemize}

    \item Dataset details (additional) (\ref{additional_dataset_section})
    
    \item Implementation details (additional)  
    (\ref{additional_implementation_section})

    \item SI-MIL additional results (\ref{additional_results})

    \item SI-MIL ablation studies: hyperparameter sensitivity (\ref{additional_ablation})

    \item SI-MIL ablation studies: model components (\ref{SI-MIL component influence})

    \item Dataset contribution details (additional) (\ref{data_contribution})
    
    \item Local interpretability analysis (additional) (\ref{additonal_interpretability})
    
    \item Global interpretability analysis (additional) (\ref{additonal_interpretability_global}) 
    
    \item Top-$K$ comparative analysis (\ref{topk_comparative_analysis}) 
    % Conventional MIL vs SI-MIL (top 20 patches with Raj's comments for 2 WSIs - BRCA)
    
    % \item Slide-level classification performance: p-value analysis (\ref{p_value})  \textcolor{red}{- \textbf{to do}}
    
    \item Hand-crafted PathExpert feature extraction (\ref{path_features}) 
    % brief description of feature group
    % for each group, describe the hyperparameters used
    % for each group, the list of features
    % for a few features (that can be translated to PathExpert terminologies) show low-medium-high: visual reasoning of features

\end{itemize}

\section{Dataset details (additional)}
\label{additional_dataset_section}

We benchmark our SI-MIL on three WSI datasets, namely \textbf{TCGA-BRCA}, \textbf{TCGA-Lung}, and \textbf{TCGA-CRC}. 
% SI-MIL build on PathExpert features extracted from WSI patches, and several of these features depend on predictions from HoVer-Net, that is trained on 40$\times$. 
SI-MIL necessitates PathExpert features for interpretable prediction. We use HoVer-Net for  segmenting and classifying the nuclei, and afterwards computed hand-crafted PathExpert features. Since HoVer-Net is trained exclusively on 40$\times$ magnification patches, our analysis is confined to WSIs having 40$\times$ magnification. This ensures accurate nuclei prediction and thereby meaningful PathExpert feature extraction.

%Since PathExpert features depends on prediction from HoVer-Net, which is trained on 40$\times$, therefore for one-to-one comparison with deep features, only the slides with 40$\times$ magnification available are selected across datasets. 
\textbf{TCGA-BRCA} is split into 825 training (653 IDC, 172 ILC) and 85 testing (67 IDC, 18 ILC) WSIs following~\cite{chen2022scaling}. \textbf{TCGA-Lung} dataset is split into 744 training (388 LUAD, 356 LUSC) and 192 testing (96 LUAD, 96 LUSC) WSIs following DSMIL~\cite{li2021dual}.
% (\href{https://github.com/binli123/dsmil-wsi}{https://github.com/binli123/dsmil-wsi}). 
For \textbf{TCGA-CRC}, following ~\cite{liu2018comparative, bilal2021development}, we use the first three folds for training, \ie, 241 WSIs (38 hypermutated, 203 not) and the fourth for testing, \ie, 79 slides (12 hypermutated, 67 not) with 40$\times$ filtering. The patch extraction process implemented in our study follows the methodology outlined in the aforementioned DSMIL repository~\cite{li2021dual}.

% should we talk about filtering with respect to number of nuclei and percentage of background? because all our features are based on the nuclei. 

% Talk about patch extraction

\section{Implementation details (additional)}
\label{additional_implementation_section}

\subsection{Deep feature extractors}
We compared SI-MIL against different baselines that include training Additive ABMIL using different types of patch features. Details of the patch feature extractors are presented as follows:

\noindent \textbf{IN ViT-S:} We train the Additive ABMIL using features extracted by a popular ImageNet-supervised model, specifically ViT-Small~\cite{dosovitskiy2020image} model pre-trained using ImageNet dataset~\cite{deng2009imagenet}. The model extracts a feature embedding of size \(D = 384\) for each WSI patch.

%We compared SI-MIL against the widely used method in computational pathology, \ie, using an ImageNet-supervised model (here we adopted IN ViT-S, \ie, ImageNet-supervised ViT-Small~\cite{dosovitskiy2020image}) directly as a feature extractor for WSI patches. This model extracts a feature embedding of size \(D = 384\) for each WSI patch.

\noindent \textbf{RetCCL:} We adopted a state-of-the-art feature extractor~\cite{wang2023retccl} pre-trained using pathology images. This benchmarks our trained feature extractor, described in Sec.~\ref{deep_baselines}. This model extracts a feature embedding of size \(D = 2048\).

\noindent \textbf{CTransPath:} Similar to RetCCL, we benchmarked against a Transformer-based feature extractor pre-trained using pathology images~\cite{wang2022transformer}. The resulting patch embeddings are of size \(D = 768\).

It is important to note that CTransPath and RetCCL were pre-trained on the pan-TCGA~\cite{tcga} dataset, and our evaluated datasets are subset of this dataset. Therefore, these models were pre-trained using the WSIs in our test dataset, which can potentially result in inflated performances during classification. Though benchmarked, these models may not be suitable for reliable comparisons in our study.

\noindent \textbf{DINO ViT-S:} For a reliable comparison, 
%To address the problem identified with using the above two pathology pre-trained feature extractors, 
we used DINO~\cite{caron2021emerging} to pre-train ViT-Small models for each dataset (TCGA-BRCA, TCGA-Lung, and TCGA-CRC), using the dataset-specific training splits as provided in the Supp Sec.~\ref{additional_dataset_section}
For pre-training, we used the default hyperparameter values of DINO~\cite{caron2021emerging}, while using only two global crops. These pre-trained models extract a feature embedding of size \(D = 384\) for each WSI patch. One RTX 8000 GPU is utilized for pretraining the ViT-S with a batch size of 256.

\subsection{SI-MIL}
\label{si_mil_supp}

Hyperparameter tuning is performed with a range of learning rates $\in \{1e^{-3}, 2e^{-3}, 1e^{-4}, 2e^{-4}\}$ and weight decays $\in \{1e^{-2}, 5e^{-3}\}$. By default, $\#$PF-Mixer layers $=$4, $\lambda = 20$, $K = 20$, $\gamma=0.75$, and $t=3$. Additionally, $d=246$ except for in TCGA-Lung where $d=203$ as annotations for only 4 (instead of 5) cell types are available for HoVer-Net classification in the Lung dataset.  For both the predictors $L(\cdot)$ and $C(\cdot)$, we use the sigmoid activation ($\psi$), since our tasks involve binary classification. Note that $\mathcal{L}_{KD}$ is utilized with stop-gradient since the goal is to align the performance of the \textit{Self-Interpretable} branch to be close to high performing conventional MIL branch in SI-MIL. All MIL experiments are performed on one RTX 8000 GPU.

\subsection{Interpretability analysis setup}

For interpretability analysis, we compare the separability of the top \( K \) patches in the PathExpert feature space between the conventional MIL and SI-MIL (refer to Figure~\ref{fig:cohort_tsne}). To ensure a fair comparison, we select WSIs from the held-out test set where both MIL methods result in correct predictions. 

Note that, we employ 5-fold cross-validation on the training split and held out the test set. We chose the best-performing fold for both local (visualization and pathologist relevancy score experiment) and global (visualization) interpretability analysis for the MIL methods.
However, for multivariate class-separability scores (refer to Figure~\ref{fig:cohort_tsne}), we report the median and standard deviation from all 5-folds. Similarly, we report the median and standard deviation of Jensen-Shannon (JS) divergence across all 5-folds in Figure~\ref{fig:cohort_tsne}.

\subsection{SI-MIL complexity analysis}
The mitigation of the trade-off between performance and interpretability by SI-MIL can be attributed to the choice of PathExpert features and the SI-MIL design choices, instead of merely an increase in the number of model parameters. It can be justified by comparing the size and performance of SI-MIL with the competing baselines. 
The number of model parameters in SI-MIL is 625K, while those in conventional MIL with DINO/IN ViT-S, CTransPath and ReTCCL are 345K, 985K, and 5.25M, respectively.
Despite the differences in model sizes, SI-MIL results in comparable performance with respect to the competing baselines, as shown in Table~\ref{table:results}.

\section{SI-MIL additional results} % sensitivity
\label{additional_results}

Here we provide the mean and standard deviations for the main experiments (refer to Table~\ref{table:results}) in Table~\ref{table:results_additional}.

\begin{table*}[h]
\caption{Results indicate the mean and standard deviation of 5-fold cross-validation on test set. All methods are trained with Additive ABMIL as base MIL. Int. denotes self-interpretability of a method.}
% \vspace{-0.4cm}
\label{table:results_additional}
% \tiny
\begin{center}
% \resizebox{\columnwidth}{!}{
\begin{tabular}{cccccccc}
\toprule
        &      & \multicolumn{2}{c}{\textbf{Lung}} & \multicolumn{2}{c}{\textbf{BRCA}} & \multicolumn{2}{c}{\textbf{CRC}}        \\ 
          & Int.    & \multicolumn{1}{c}{Acc.}       & AUC    & \multicolumn{1}{c}{Acc.}       & AUC   & \multicolumn{1}{c}{Acc.}       & AUC     \\  
\midrule

% IN ViT-S    & \xmark         &    0.859 \pm 0.014  &    0.919 \pm 0.004   &    0.929 \pm 0.011  &    0.967 \pm 0.005  &    0.891 \pm 0.013 &    0.898 \pm 0.018  \\ 

IN ViT-S    & \xmark         &    $0.859 \pm 0.014$  &    $0.919 \pm 0.004$   &    $0.929 \pm 0.011$  &    $0.967 \pm 0.005$  &    $0.891 \pm 0.013$ &    $0.898 \pm 0.018$  \\ 

   % RetCCL          & \xmark     &      0.860 \pm 0.008 &    0.935 \pm 0.003  &    0.929 \pm 0.011 &    \textbf{0.976} \pm \textbf{0.001}  &    0.889 \pm 0.015  &    0.891 \pm 0.047   \\ 

  RetCCL          & \xmark     &      $0.860 \pm 0.008$ &    $0.935 \pm 0.003$  &    $0.929 \pm 0.011$ &    $\textbf{0.976} \pm \textbf{0.001}$  &    $0.889 \pm 0.015$  &    $0.891 \pm 0.047$   \\

  % CTransPath          & \xmark     &      \textbf{0.904}  \pm \textbf{0.003} &    \textbf{0.967}  \pm \textbf{0.002}  &    0.920 \pm 0.023 &    0.974  \pm 0.002 &    \textbf{0.906} \pm \textbf{0.010} &    \textbf{0.897} \pm \textbf{0.023} \\ 

  CTransPath          & \xmark     &      $\textbf{0.904}  \pm \textbf{0.003}$ &    $\textbf{0.967}  \pm \textbf{0.002}$  &    $0.920 \pm 0.023$ &    $0.974  \pm 0.002$ &    $\textbf{0.906} \pm \textbf{0.010}$ &    $\textbf{0.897} \pm \textbf{0.023}$ \\ 

  % DINO ViT-S     & \xmark   &      0.896 \pm 0.003 &    0.957 \pm0.003  &    \textbf{0.937} \pm \textbf{0.012} &   0.974 \pm 0.005  &    0.904 \pm 0.006 & \textbf{0.897} \pm \textbf{0.014}  \\ 

  DINO ViT-S     & \xmark   &      $0.896 \pm 0.003$ &    $0.957 \pm0.003$  &    $\textbf{0.937} \pm \textbf{0.012}$ &   $0.974 \pm 0.005$  &    $0.904 \pm 0.006$ & $\textbf{0.897} \pm \textbf{0.014}$  \\

\midrule

% PathFeat       & \xmark    &      0.830 \pm 0.015  &    0.888 \pm 0.009   &    0.885 \pm 0.014  &    0.950 \pm 0.005  &    \textbf{0.886} \pm \textbf{0.016}  &    0.818 \pm 0.031    \\ 

PathFeat       & \xmark    &      $0.830 \pm 0.015$  &    $0.888 \pm 0.009$   &    $0.885 \pm 0.014$  &    $0.950 \pm 0.005$  &    $\textbf{0.886} \pm \textbf{0.016}$  &    $0.818 \pm 0.031$    \\

 % PathFeat w/o $H(\cdot)$        & \cmark    &      0.767 \pm 0.018  &    0.837 \pm 0.016  &    0.889 \pm 0.012  &    0.914 \pm 0.003  &    0.853  \pm 0.013 &    0.720 \pm 0.044  \\ 

 PathFeat w/o $H(\cdot)$        & \cmark    &      $0.767 \pm 0.018$  &    $0.837 \pm 0.016$  &    $0.889 \pm 0.012$  &    $0.914 \pm 0.003$  &    $0.853  \pm 0.013$ &    $0.720 \pm 0.044$  \\

 % 2-stage training & \cmark  &      0.865 \pm 0.007  &    0.932 \pm 0.009  &    0.908 \pm 0.017 &    0.924 \pm 0.019  &    0.876 \pm 0.020   &    0.862 \pm 0.036  \\ 

 2-stage training & \cmark  &      $0.865 \pm 0.007$  &    $0.932 \pm 0.009$  &    $0.908 \pm 0.017$ &    $0.924 \pm 0.019$  &    $0.876 \pm 0.020$   &    $0.862 \pm 0.036$  \\

   % SI-MIL (ours)   & \cmark   &    \textbf{0.884} \pm \textbf{0.018}  &    \textbf{0.941} \pm \textbf{0.009} &    \textbf{0.944} \pm \textbf{0.028} &    \textbf{0.968} \pm \textbf{0.012}  &    0.884 \pm 0.017   &    \textbf{0.910} \pm \textbf{0.018}   \\ 
   
SI-MIL (ours)   & \cmark   &    $\textbf{0.884} \pm \textbf{0.018}$  &    $\textbf{0.941} \pm \textbf{0.009}$ &    $\textbf{0.944} \pm \textbf{0.028}$ &    $\textbf{0.968} \pm \textbf{0.012}$  &    $0.884 \pm 0.017$   &    $\textbf{0.910} \pm \textbf{0.018}$   \\ 

% \midrule
\midrule

 \multicolumn{8}{c}{\textbf{Ablation study of SI-MIL components}} \\
\midrule

 % w/o  PAG Top-$K$  & \cmark  &     0.859 \pm 0.009 &    0.936  \pm 0.011 &    0.915 \pm 0.023 &    0.922 \pm 0.026  &    0.876 \pm 0.022   &    0.869 \pm 0.024   \\ 

 %  w/o  KD   & \cmark   &    0.853 \pm 0.010  &    0.915 \pm 0.007  &    0.932 \pm 0.016  &    0.951 \pm 0.024   &    0.878 \pm 0.024 &    0.830 \pm 0.039    \\ 

 % w/o PAG Top-$K$ \& KD & \cmark  &      0.857 \pm 0.005 &    0.924 \pm 0.005  &    0.915 \pm 0.009 &    0.899 \pm 0.013  &    0.879  \pm 0.022  &    0.858 \pm 0.036  \\ 

w/o PAG Top-$K$  & \cmark  &     $0.859 \pm 0.009$ &    $0.936  \pm 0.011$ &    $0.915 \pm 0.023$ &    $0.922 \pm 0.026$  &    $0.876 \pm 0.022$   &    $0.869 \pm 0.024$   \\ 

w/o KD   & \cmark   &    $0.853 \pm 0.010$  &    $0.915 \pm 0.007$  &    $0.932 \pm 0.016$  &    $0.951 \pm 0.024$   &    $0.878 \pm 0.024$ &    $0.830 \pm 0.039$    \\ 

w/o PAG Top-$K$ \& KD & \cmark  &      $0.857 \pm 0.005$ &    $0.924 \pm 0.005$  &    $0.915 \pm 0.009$ &    $0.899 \pm 0.013$  &    $0.879  \pm 0.022$  &    $0.858 \pm 0.036$  \\

\bottomrule

\end{tabular}
% }
\end{center}
% \vspace{-0.8cm}
\end{table*}

\section{SI-MIL ablation studies: hyperparameters} % sensitivity
\label{additional_ablation}

In this section, we provide studies of SI-MIL hyperparameters on TCGA-BRCA dataset. Particularly, these ablations demonstrate the effect of varying \( K \) in the PAG Top-$K$ module, the number of PF-Mixer layers, and the percentile and temperature for scaling \( \beta \).

\begin{figure}[h!]
\centering
    \includegraphics[width=1\linewidth]{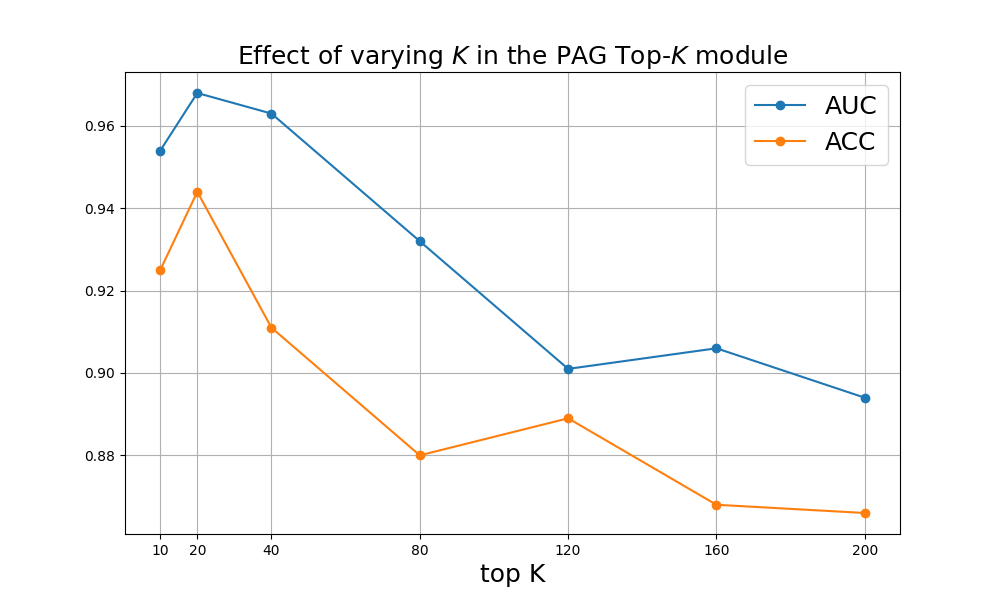}
    \caption{PAG Top-$K$ module ablation}
    \label{fig:topK}
\end{figure}

\textbf{Effect of varying $K$ in the PAG Top-$K$ module:}
In Figure~\ref{fig:topK}, we illustrate the impact of varying $K$ on SI-MIL performance. We observe that a larger value of $K$ leads to a significant drop in performance compared to the default $K=20$. This decrease may be attributed to an increase in irrelevant noisy patches, which makes it difficult for the model to classify WSIs in the PathExpert feature space.

\begin{figure}[h!]
\centering
    \includegraphics[width=1\linewidth]{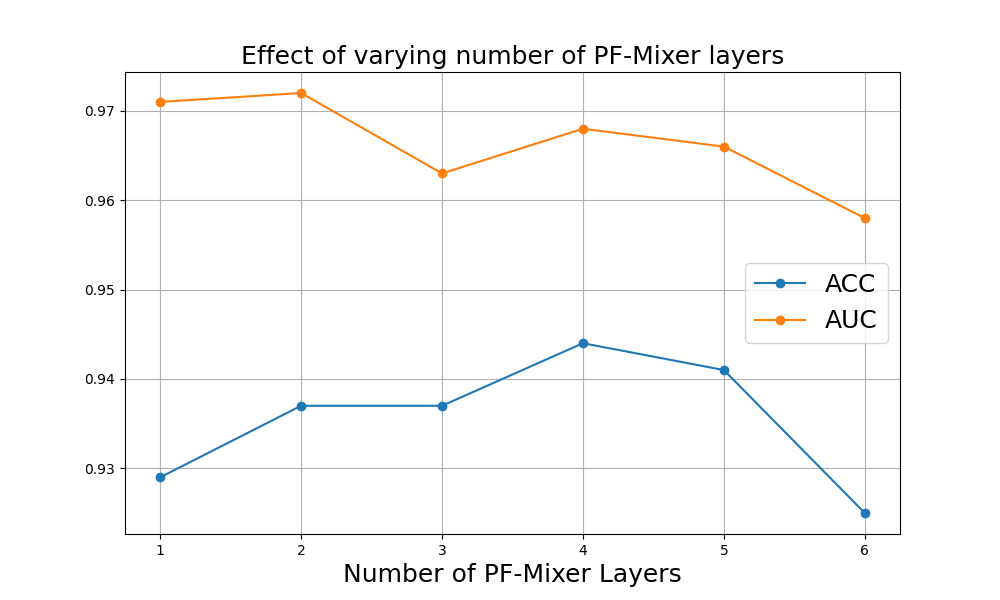}
    \caption{PF-Mixer module layers ablation}
    \label{fig:pfmixer}
\end{figure}

\noindent \textbf{Effect of varying number of PF-Mixer layers:} 
SI-MIL's performance is generally robust across various values of the number of PF-Mixer layers, but experiences a performance drop for very high values, \eg, \#PF-Mixer layers $=6$ (Figure~\ref{fig:pfmixer}). This decline can be attributed to potential overfitting induced as a result of higher number of layers.

\begin{figure}[ht!]
\centering
    \includegraphics[width=1\linewidth]{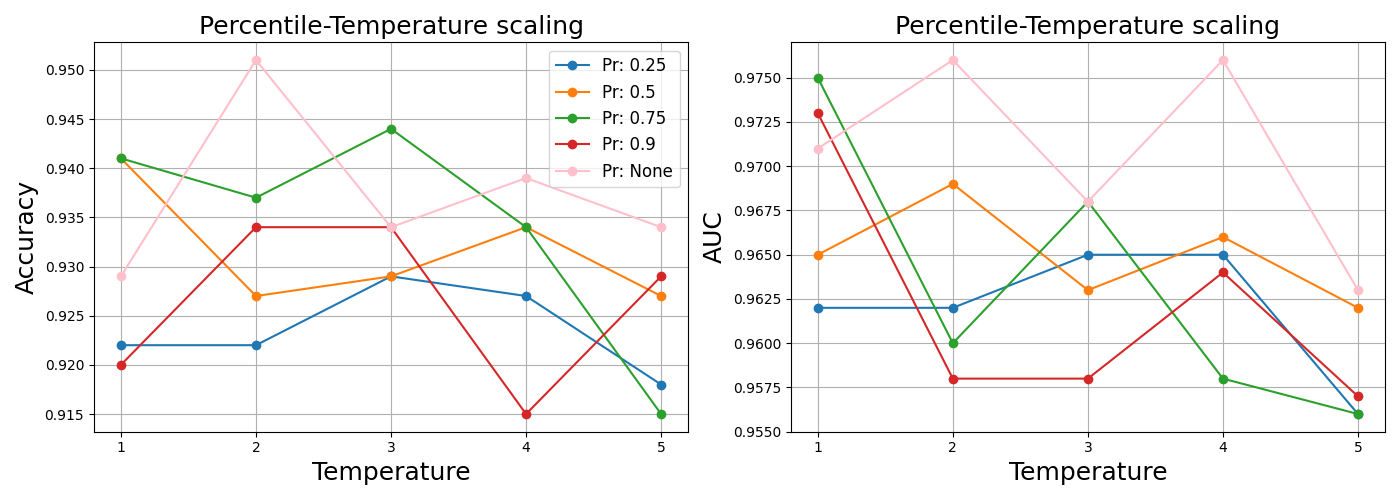}
    \caption{$\beta$ scaling ablation}
    \label{fig:scaling}
\end{figure}

\noindent \textbf{Effect of percentile and temperature for scaling $\beta$:}
In Figure~\ref{fig:scaling}, we show the variation in performance of SI-MIL with respect to the percentile value ($Pr_{\gamma}$) and temperature ($t$) for scaling the feature attention values $\beta$ in eq.~\ref{b_scale}. ``None" in Figure~\ref{fig:scaling} refers to the absence of percentile and standard deviation scaling.

$Pr_{\gamma}$ controls the percentage of features ($d$) that have a positive value before being fed to the sigmoid activation in eq.~\ref{b_scale}. The temperature parameter ($t$) determines the sharpness of this curve, with a high value indicating that most values deviate from zero before being fed to the sigmoid. Thus, having high $Pr_{\gamma}$ and high $t$ leads to a very sparse selection of features. Since our goal is to interpret the prediction of WSI, it is beneficial to explain the prediction in terms of the contribution of ``few'' most discriminative features. Note that the absence of $Pr_{\gamma}$ scaling and/or low temperature allows the model to use a large number of features for its prediction; thus making it harder to interpret the predictions. Therefore, the main goal is to have higher values of $Pr_{\gamma}$ and $t$, while maintaining a good SI-MIL performance.

In Figure~\ref{fig:scaling}, we can observe that having no $Pr_{\gamma}$ scaling generally results in the best performance, whereas a very high value, such as $Pr_{\gamma} = 0.9$, performs poorly in most cases. We find that having a slightly lower $Pr_{\gamma} = 0.75$ and $t = 3$ establishes an optimal balance, by enforcing adequate sparsity while still performing efficiently.

\section{SI-MIL ablation studies: components}
\label{SI-MIL component influence}

In Table~\ref{table:results}, we demonstrate the variations in performance when ablating different components of SI-MIL. Similarly, in Table~\ref{table:SI-MIL ablation}, we present additional experiments on the impact of different SI-MIL components.

We observed that omitting the \textit{Feature Attention module} $A^f(\cdot)$ results in better performance compared to using it without the \textit{PF-Mixer} network ${Mix}(\cdot)$, though both scenarios underperform relative to the proposed SI-MIL. This indicates that $A^f(\cdot)$, which softly selects features, requires contextualization among the patches and features before highlighting or attenuating specific features within this module. Without appropriate contextualization, processing each feature row in matrix $M^T$ independently leads to suboptimal decisions by $A^f(\cdot)$ and reduces performance.

We further investigate the necessity of deep features in SI-MIL. For this purpose, we substituted deep features with PathExpert features in the conventional MIL branch, thereby using the same PathExpert features in both SI-MIL branches. As shown in Table~\ref{table:SI-MIL ablation}, the performance declines with or without $\mathcal{L}_{KD}$ when replacing deep features, underscoring the importance of employing potent deep features to guide the \textit{Self-Interpretable} branch in SI-MIL.

\begin{table}[ht]
\caption{Results indicate the mean of 5-fold cross-validation on test set. All methods are trained with Additive ABMIL as the base MIL. Int. denotes self-interpretability of a method.}
\label{table:SI-MIL ablation}
\small
\begin{center}
\resizebox{\columnwidth}{!}{
\begin{tabular}{cccccccc}
\toprule
         &      & \multicolumn{2}{c}{\textbf{Lung}} & \multicolumn{2}{c}{\textbf{BRCA}} & \multicolumn{2}{c}{\textbf{CRC}}        \\ 
          & Int.    & \multicolumn{1}{c}{Acc.}       & AUC    & \multicolumn{1}{c}{Acc.}       & AUC   & \multicolumn{1}{c}{Acc.}       & AUC     \\  

\midrule

PathFeat       & \xmark    &      0.830  &    0.888   &    0.885  &    0.950   &    \textbf{0.886}  &    0.818     \\ 

 PathFeat w/o $H(\cdot)$        & \cmark    &      0.767   &    0.837   &    0.889  &    0.914   &    0.853   &    0.720   \\ 

 2-stage training & \cmark  &      0.865  &    0.932   &    0.908  &    0.924   &    0.876    &    0.862   \\

 SI-MIL (ours)  & \cmark   &    \textbf{0.884}   &    \textbf{0.941}   &    \textbf{0.944}  &    0.968   &    0.884    &    \textbf{0.910}    \\ 

% \addlinespace[1pt]
% \cdashline{1-8}
% \addlinespace[2pt]

\midrule

 \multicolumn{8}{c}{\textbf{Ablation study of SI-MIL components}} \\
\midrule

% \multirow{6}{*}{\makecell{SI-MIL \\ Ablations}} 
 % wo/  Pert Top-K  & \cmark  &     0.859  &    0.936   &    0.915  &    0.922   &    0.876    &    0.869    \\ 

w/o  $A^f(\cdot)$     & \cmark   & 0.853  &    0.935   &    0.939  &    \textbf{0.981}   &    0.871  &    0.857    \\ 

  w/o  ${Mix}(\cdot)$     & \cmark   & 0.838  &    0.915   &    0.925  &    0.953   &    0.866  &    0.863    \\

  w/  PathFeat only     & \cmark   & 0.863  &    0.936   &    0.911  &    0.942   &    0.876  &    0.836    \\

  w/  PathFeat only \&  w/o KD     & \cmark   & 0.847  &    0.911   &    0.911  &    0.945   &    0.853  &    0.781    \\

 %  wo/  KD   & \cmark   &    0.853   &    0.915   &    0.932  &    0.951   &    0.878  &    0.830     \\ 

 % wo/ Pert Top-K wo/ KD & \cmark  &      0.857  &    0.924   &    0.915  &    0.899   &    0.879    &    0.858   \\ 

\bottomrule

\end{tabular}
}
\end{center}
\end{table}

\section{Dataset contribution details (additional)}
\label{data_contribution}

We contribute a unique comprehensive dataset aimed at enhancing interpretability and reproducibility in MIL research. The dataset encompasses nuclei maps and PathExpert features for over 2,200 WSIs. SI-MIL-generated patch-feature importance reports will also be made available for representative slides. It covers multiple organs and cancer types, including Lung (lung adenocarcinoma vs. lung squamous cell carcinoma), Breast (invasive ductal vs. invasive lobular carcinoma), and Colon (low vs. high mutation). This diverse collection facilitates in-depth studies across various cancer types, providing a valuable resource for advancements in interpretable MIL methodologies.

Successful translation of AI tools to the clinic hinges upon the interpretability and trustworthiness of the tools. This dataset will serve as a critical asset for both the medical vision and digital pathology communities, facilitating the exploration of new research directions in the development of interpretable AI techniques for computational pathology. A significant obstacle in digital pathology research has been the intensive resource requirements for extracting features that possess clear geometric and physical significance, and which are interpretable by pathologists. The dataset creation involved analyzing gigapixel WSIs at 40$\times$ magnification, leveraging HoVer-Net~\cite{graham2019hover} for cell segmentation and classification, followed by extracting PathExpert features and feature importance scores detailed in Sec.~\ref{feature_extraction} and Sec.~\ref{sec_intepretability}, respectively. 
Processing each WSI required $\sim$2 hours, divided between GPU-based cell map prediction and CPU-based PathExpert feature extraction. Employing three RTX 8000 GPUs and a 40-core CPU with 500GB RAM, the total processing amounted to $\sim$4400 hours ($\approx$60 days).
We provide the comprehensive set to enable further research.

In view of $\sim$2 TB memory foorprint of HoVer-Net nuclei maps and the processed PathExpert features, we intend to host this dataset on TCIA Analysis Results, akin to other popular preprocessed datasets~\cite{kalpathy10crowds, saltz2018tumor}.

% \href{https://wiki.cancerimagingarchive.net/pages/viewpage.action?pageId=33948774}{https://tinyurl.com/mr3fnuhz} and \href{https://wiki.cancerimagingarchive.net/pages/viewpage.action?pageId=33948919}{https://tinyurl.com/2j4ty3vm}.

% This dataset will be released under the Creative Commons Attribution 4.0 International License (CC BY 4.0), which allows for sharing, copying, and redistribution in any medium or format, as well as adaptation, remixing, transforming, and building upon the material for any purpose, even commercially, as long as appropriate credit is given, a link to the license is provided, and any changes made are indicated. The full details of the license can be accessed at \href{https://creativecommons.org/licenses/by/4.0/}{https://creativecommons.org/licenses/by/4.0/}.

The dataset will be released under the Creative Commons Attribution-NonCommercial 4.0 International License (\href{https://creativecommons.org/licenses/by-nc/4.0/}{CC BY-NC 4.0}). It permits the sharing, copying, and redistribution in any medium or format, as well as adaptation, remixing, transforming, and building upon the material for non-commercial purposes. Appropriate credit must be given, a link to the license must be provided, and any changes made should be indicated.

% \pp{Can we add a few sentences on how this dataset can be used by digital pathology researchers - how it can advance the field? Also, how do we plan to maintain it/update it?}

\section{Local interpretability analysis (additional)}
\label{additonal_interpretability}

\begin{figure*}[ht!]
\centering
    \includegraphics[width=1\linewidth]{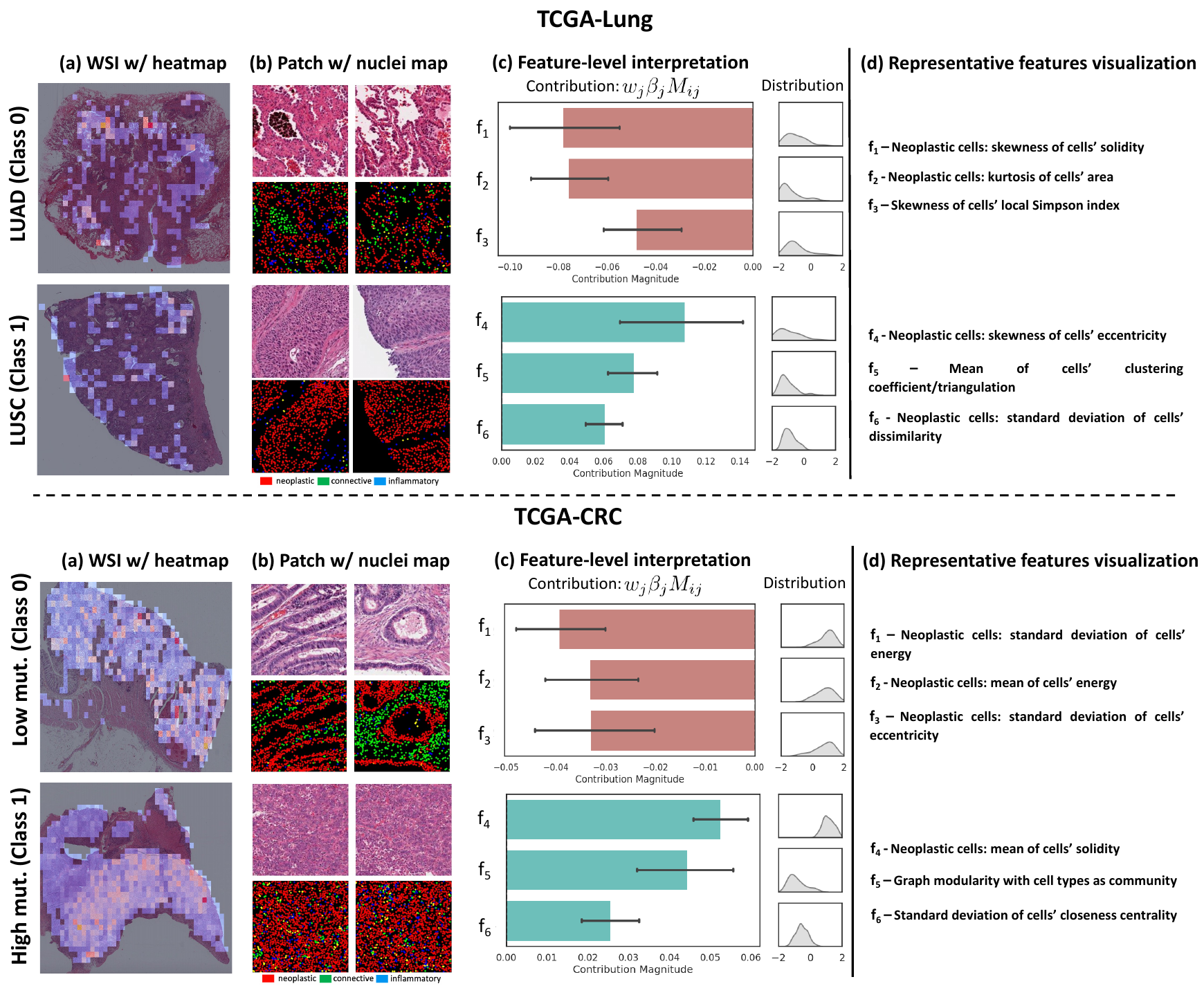}
    \caption{\textbf{Qualitative Patch-Feature importance report:} In (a) and (b), we present WSIs with overlaid attention heatmaps and the top two patches, along with their nuclei maps. In (c), we demonstrate the mean contribution magnitude of select representative features across the top $K$ patches employed in the Self-Interpretable branch. Additionally, we display a feature density plot that quantifies the distribution of features within the $K$ patches. For brevity, we omit the y-axis. Given that these features are normalized, a curve leaning towards the right indicates higher/positive values, while one towards the left signifies lower/negative values, depending on the feature. Finally, in (d), we illustrate, the description of representative features in (c).}
    \label{fig:patient_interpretable_lung_crc}
\end{figure*}

Here, we present additional predictions (refer to Sec.~\ref{sec_intepretability}) for WSIs from other datasets, \ie, TCGA-Lung and TCGA-CRC. Please note that the predictions for all WSIs in the evaluated datasets will be released as part of the contributed dataset. Qualitative patch-feature importance reports for TCGA-Lung and TCGA-CRC are illustrated in the upper and lower half of Figure~\ref{fig:patient_interpretable_lung_crc}, respectively.

% \begin{figure*}[ht!]
% \centering
%     \includegraphics[width=1\linewidth]{figures/Patient-level_interpretation crc.pdf}
%     \caption{\textbf{Qualitative Patch-Feature importance report (TCGA-CRC):} In (a) and (b), we present WSIs with overlaid attention heatmaps and the top two patches, along with their nuclei maps. In (c), we demonstrate the mean contribution magnitude of select representative features across the top $K$ patches employed in the Self-Interpretable branch. Additionally, we display a feature density plot that quantifies the distribution of features within the $K$ patches. For brevity, we omit the y-axis. Given that these features are normalized, a curve leaning towards the right indicates higher/positive values, while one towards the left signifies lower/negative values, depending on the feature. Finally, in (d), we illustrate, the description of representative features in (c).}
%     \label{fig:patient_interpretable_crc}
% \end{figure*}

\section{Global interpretability analysis (addition)}
\label{additonal_interpretability_global}

\begin{figure*}[ht!]
\centering
    \includegraphics[width=1\linewidth]{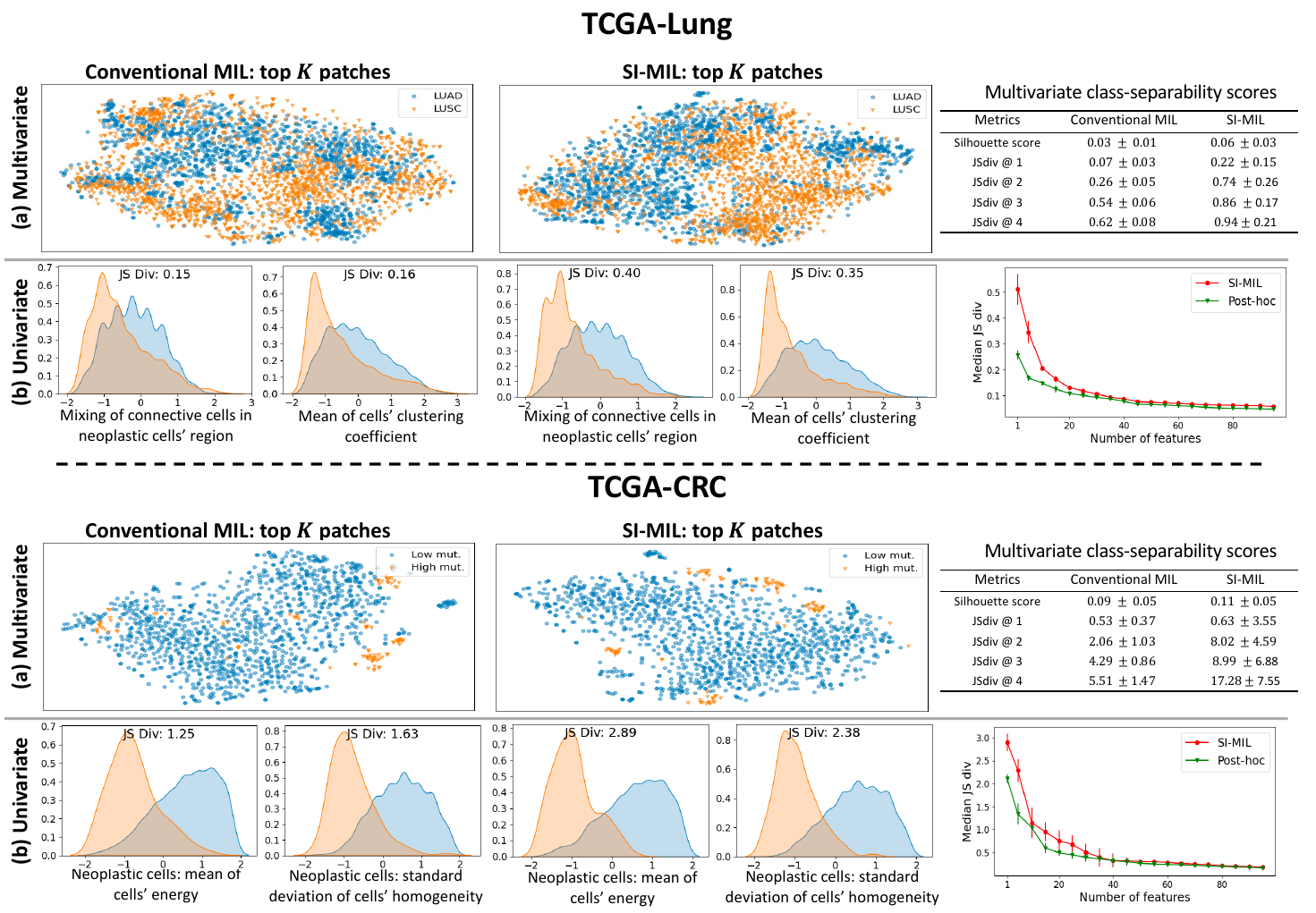}
    \caption{\textbf{Cohort-level Interpretation:} Separability of top $K$ patches of WSIs across classes in the PathExpert feature space. Multivariate and Univariate analyses depict that the top $K$ patches selected by SI-MIL and their PathExpert features are more separable.}
    \label{fig:cohort_tsne_lung_crc}
\end{figure*}

Here, we present global interpretability analysis (refer to Sec.~\ref{sec_intepretability}) for patches from the test set WSIs. We include only the WSIs that were correctly predicted by both the conventional MIL and SI-MIL, to ensures a fair comparison, as described in Sec.~\ref{additional_implementation_section}. Cohort-level interpretation for TCGA-Lung and TCGA-CRC are illustrated in the upper and lower half of Figure~\ref{fig:cohort_tsne_lung_crc}, respectively.

% \begin{figure*}[ht!]
% \centering
%     \includegraphics[width=1\linewidth]{figures/Cohort_tsne_brca_plot_crc.pdf}
%     \vspace{-0.3cm}
%     \caption{\textbf{Cohort-level Interpretation (TCGA-CRC):} Separability of top $K$ patches of WSIs across classes in the PathExpert feature space. Multivariate and Univariate analyses depict that the top $K$ patches selected by SI-MIL and their PathExpert features are more separable.}
%     \vspace{-0.3cm}
%     \label{fig:cohort_tsne_crc}
% \end{figure*}

\section{Top-$K$ comparative analysis}
\label{topk_comparative_analysis}

% Analysis with Raj here about topk changing between few slides.

\begin{figure*}[p]
\centering
    \includegraphics[width=1\linewidth]{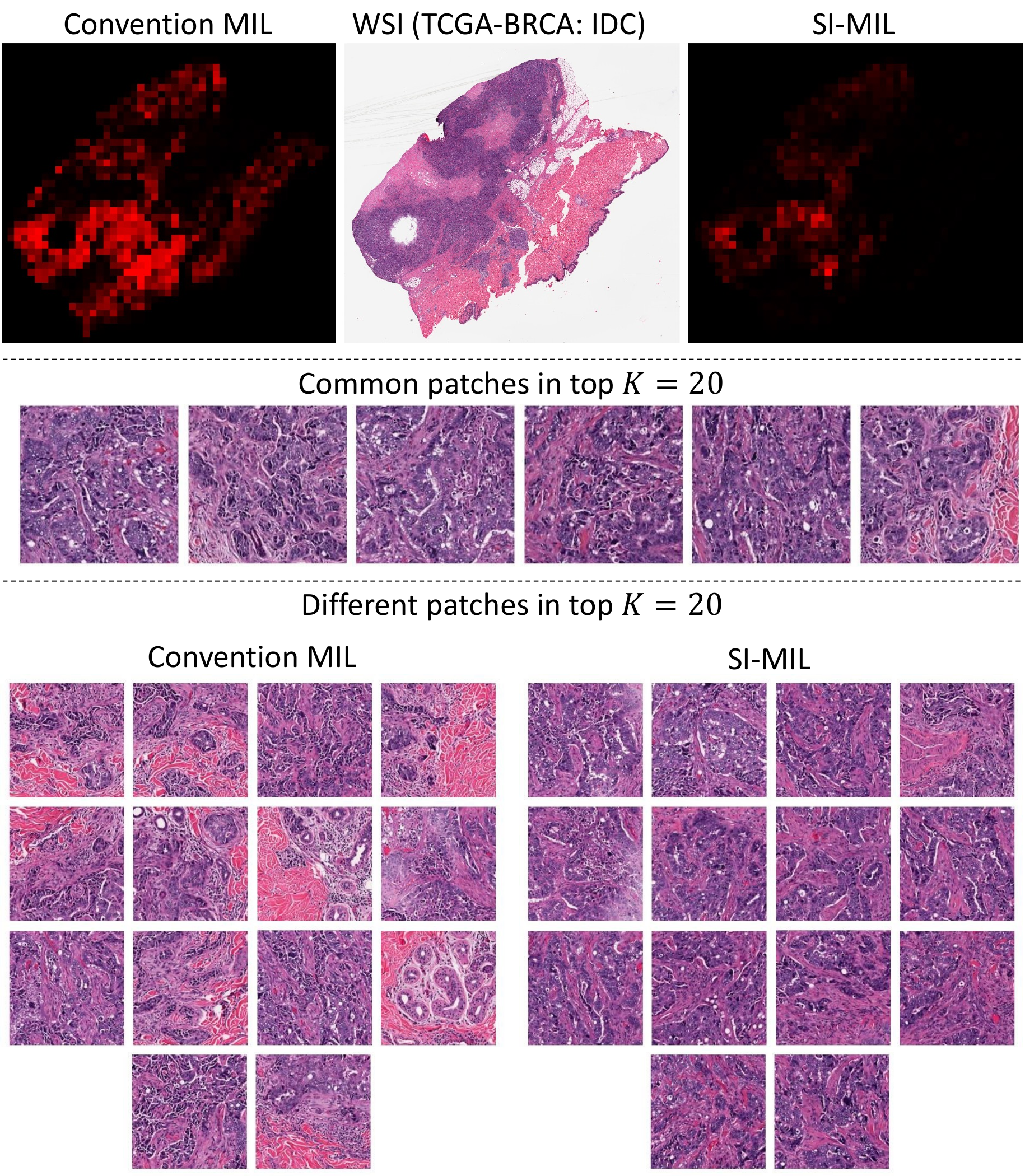}
    \caption{TCGA-BRCA Invasive Ductal Carcinoma (IDC) sample. Refinement of the patch attention map by the Self-Interpretable branch, transitioning from conventional MIL to SI-MIL.}
    \label{fig:topk_analysis_brca_idc}
\end{figure*}

\begin{figure*}[p]
\centering
    \includegraphics[width=1\linewidth]{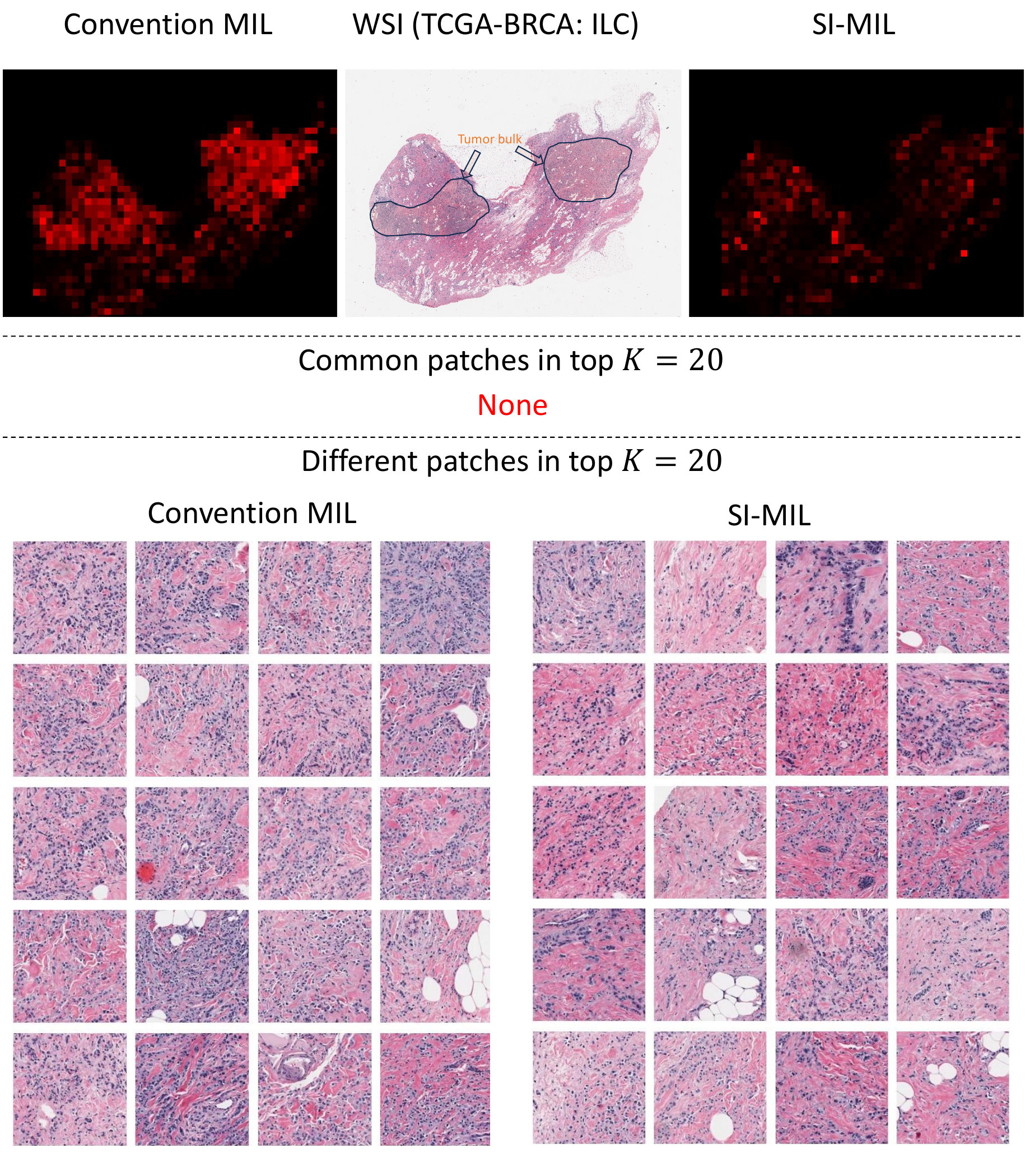}
    \caption{TCGA-BRCA Invasive Lobular Carcinoma (ILC) sample. Refinement of the patch attention map by the Self-Interpretable branch, transitioning from conventional MIL to SI-MIL.}
    \label{fig:topk_analysis_brca_ilc}
\end{figure*}

In this section, we demonstrate how the \textit{Self-Interpretable} branch of SI-MIL tames the patch attention map of conventional deep MIL. Specifically, Figures~\ref{fig:topk_analysis_brca_idc} and~\ref{fig:topk_analysis_brca_ilc} compare the spatial attention maps generated after the training of conventional MIL (\ie, without PathExpert features) and SI-MIL, which integrates both the conventional MIL and \textit{Self-Interpretable} branches. We proceed to visualize the top $K=20$ patches from both MIL methods, categorizing them into groups based on whether they are common or distinct between the methods.

In Figure~\ref{fig:topk_analysis_brca_idc}, we contrast the top \(K=20\) patch selection of our Self-Interpretable MIL (SI-MIL) with conventional MIL in analyzing invasive ductal carcinoma (IDC) WSIs. SI-MIL and conventional MIL share 6 out of 20 patches, but differ in the remaining 14. \uline{While conventional MIL often chooses patches near the dermis, featuring IDC with smooth connective areas and occasionally normal glands, SI-MIL targets patches indicative of malignancy, marked by malignant cancerous ducts with large, distorted nuclei.} This difference, especially evident in the unique patches of SI-MIL, underscores its focus on diagnostically relevant areas like malignant glands with compressed lumens and hyperchromatic nuclei, contrasted against the tissue highlighted by conventional MIL. SI-MIL's emphasis is on patches comprising 70-80\% of malignant features, including dense pink-colored cancer-associated stroma, aligning with its goal of accurate diagnosis.

In the context of invasive lobular carcinoma (ILC), early detection is crucial due to its rapid spread and poor long-term survival outcomes, necessitating clear differentiation from invasive ductal carcinoma (IDC). In Figure~\ref{fig:topk_analysis_brca_ilc}, we observe an absence of common patches between the top \(K=20\) attended patches of both MIL methods. 
\uline{Our method, in contrast to conventional MIL, distinctively focuses on invasive single file chains, often found at the periphery of the tumor bulk or the invasive front, which are more characteristic of ILC. This is in contrast to the conventional MIL's emphasis on patches with high cellularity within the tumor bulk.} 
The rapid spread of lobular cancer is evident in its infiltration through various tissues, and unlike IDC, which often presents as glandular structures with clear separations between tumorous and connective nuclei, ILC is characterized by discohesive arrangements, leading to single file patterns with a notable mixing of tumor nuclei with connective nuclei.

% The advanced-stage presentation due to vague clinical findings, such as thickened breast tissue without a distinct mass and often ill-defined borders. 

% In Figure~\ref{fig:comparison}, we observe no common patches between the top-20 attended patches of both MIL methods. Conventional MIL typically focuses on the tumor bulk, while our SI-MIL targets the tumor periphery or invasive front. This is evident in the unique patches where conventional MIL features high tumor fraction, in contrast to the mixed tumor and connective/stroma cells in SI-MIL patches. Since such local mixing is not prevalent in IDC, SI-MIL's focus on these areas is crucial for accurately classifying IDC vs. ILC, a task challenging due to ILC's subtle presentation and lack of common necrosis.

\section{Hand-crafted PathExpert feature extraction}
\label{path_features}

In Sec.~\ref{feature_extraction}, we briefly discussed the categories of handcrafted PathExpert features, such as \textit{Morphometric} and \textit{Spatial distribution} properties. This section provides a detailed description of these features, accompanied by visualizations to elucidate the significance of their geometrical and physical meaning in computational pathology. We further refine these features into three categories, based on the studies from which they were adopted: \underline{\textit{Morphometric properties}} (from FLocK~\cite{lu2021feature}), \underline{\textit{Graph-based Social Network Analysis}}~\cite{zamanitajeddin2021cells}, and \underline{\textit{Spatial heterogeneity properties}}~\cite{martinelli2022athena}.

\subsection{Feature categories}

We employ HoVer-Net~\cite{graham2019hover} to segment and classify nuclei in each WSI patch $p_i$, using the model trained on PanNuke. The classified nuclei include Neoplastic epithelial, Connective, Inflammatory, Necrosis, and Non-neoplastic epithelial classes. Subsequently, image-processing tools are used to quantify the properties and spatial distribution of nuclei in each patch. Next we provide a detailed description of each feature within these categories.

\subsubsection{Morphometric properties} 

In a patch \(p_i\), we extract 10 morphometric properties for each segmented nucleus as outlined in Table~\ref{morphometric_prop}. To represent the entire patch, these nuclei-level features are aggregated using 4 statistical properties: mean, standard deviation, skewness, and kurtosis. This aggregation is performed separately for each of the 5 nuclei classes identified by HoVer-Net. Additionally, the number of nuclei in each class is included as a feature.

Consequently, this results in a total of 205 patch-level aggregated morphometric properties: \(10 \times 4 \times 5\) for the morphometric properties aggregated across the four statistics and five nuclei classes, plus five for the count of nuclei in each class.

In computational pathology, these 205 morphometric properties from each patch \(p_i\) provide a holistic tissue profile. These features encapsulate key morphological characteristics of nuclei, crucial for pathological assessment. By employing statistics like mean, standard deviation, skewness, and kurtosis, we gain insights into the variability, asymmetry, and tailedness of the nuclei's morphological properties within each patch. Separately analyzing these features for each of the five nuclei classes as identified by HoVer-Net enriches the model's understanding of the tissue heterogeneity and cellular composition. Additionally, counting nuclei per class quantifies cellular composition, further enriching the diagnostic value in computational pathology. 

% \pp{Is it possible to show a cartoon illustrating some of the the different properties?}

\begin{table}[h]
\centering
\begin{tabular}{cc}
\hline
\textbf{Group} & \textbf{Feature} \\
\hline
\multirow{4}{*}{Shape} & Area \\
                       & Eccentricity \\
                       & Roundness \\
                       & Orientation \\
\hline
\multirow{6}{*}{Morphology} & Mean of Intensity \\
                            & Standard Deviation of Intensity \\
                            & Contrast of Texture \\
                            & Dissimilarity of Texture \\
                            & Homogeneity of Texture \\
                            & Energy of Texture \\
\hline
\end{tabular}
\caption{Description of extracted morphometric properties for each segmented nuclei}
\label{morphometric_prop}
\end{table}

\subsubsection{Graph-based Social Network Analysis}

In a patch \(p_i\), we construct a graph based on the centroids of nuclei and quantify the properties of this network. Drawing inspiration from~\cite{zamanitajeddin2021cells}, we initially create a k-nearest neighbor graph (with \(k = 6\)) using the centroid locations of each segmented nucleus, irrespective of their classes. Subsequently, we extract 4 traditional social network analysis properties for each nucleus, as detailed in Table~\ref{social_network}. This is followed by statistical aggregation to patch-level using  mean, standard deviation, skew, kurtosis, and max. This results in total 20 aggregated Social Network features.

\begin{table}[h]
\centering
\begin{tabular}{c}
\hline
 \textbf{Feature} \\
\hline
  Degree \\
                        Degree centrality \\
                        Clustering coefficient \\
                        Closeness centrality \\
\hline
\end{tabular}
\caption{Description of extracted social network analysis properties for each nuclei}
\label{social_network}
\end{table}

The Degree and Degree Centrality metric in our study provides insight into the number of direct connections a nucleus has within the tissue network, illuminating its level of interaction. This is pivotal in understanding nuclei communication and behavior in various pathological states. The Clustering Coefficient is another key measure, offering insights into the extent of interconnectivity among a nucleus's neighbors. This can reveal localized nuclei clusters, a feature often observed in certain pathological conditions. Lastly, Closeness Centrality assesses the average shortest distance from a nucleus to all others, aiding in identifying nuclei that are central or isolated within the tissue architecture. This comprehensive analysis of nuclei organization and interaction patterns through these SNA features is crucial for an in-depth understanding of the tissue's pathology.

\subsubsection{Spatial Heterogeneity properties}

This feature group goes beyond analyzing just the centroids of nuclei; it also incorporates their classes to assess the spatial heterogeneity of various nuclei communities within a patch~\cite{martinelli2022athena}. Heterogeneity is quantified at both global and local levels in each patch.

\underline{Global level:} A range of entropy-based descriptors and k-function metrics are utilized, examining all segmented nuclei to evaluate the uniformity versus randomness in their spatial distribution. These global heterogeneity descriptors are listed in Table~\ref{athena_global}.

\underline{Local level:} Following the methodology in~\cite{martinelli2022athena}, we construct a k-nearest neighbor graph (with \(k = 6\)) using the nuclei. For each nucleus, entropy and interaction-based properties are extracted, focusing on immediate neighbors. A local interaction-score is then aggregated at the patch level, as per the process in~\cite{martinelli2022athena}. Additionally, skewness of entropy property distribution across nuclei is computed. This skewness metric discerns whether most nuclei have lower, medium, or higher entropy values, thus offering a detailed view of cellular interactions and complexity. This local-level approach highlights the intermixing of different nuclei communities, taking into account their spatial relationships, an aspect overlooked by global entropy-based descriptors. These local-level features are enumerated in Table~\ref{athena_local}.

 This results in total 21 Spatial Heterogeneity features (9 Global and 12 Local). For an in-depth explanation and visualization of these features, we direct readers to the seminal work by~\cite{martinelli2022athena}, which extensively explores these methodologies and their implications.

\begin{table}[h]
\centering
% \tiny
\resizebox{\columnwidth}{!}{
\begin{tabular}{cc}
\toprule
% \multicolumn{2}{c}{\textbf{Global Spatial Heterogeneity Descriptor}} \\
\textbf{Group} & \textbf{Feature} \\
\midrule
\multirow{5}{*}{Global Entropy} 
 &Global Shannon index \\
 &Global Simpson index \\
 &Global max entropy \\
 &Global Richness (number of cell-types present) \\
 & Graph modularity with cell types as community \\
\midrule
\multirow{4}{*}{k-function}
 &Neoplastic cells: k-function at radius 224 pixels \\
 &Neoplastic cells: k-function at radius 448 pixels \\
 &Neoplastic cells: k-function at radius 672 pixels \\
 &Neoplastic cells: k-function at radius 896 pixels \\
\bottomrule
\end{tabular}
}
\caption{Global Spatial Heterogeneity Descriptors}
\label{athena_global}
\end{table}

\begin{table}[h]
\centering
% \tiny
\resizebox{\columnwidth}{!}{
\begin{tabular}{cc}
\toprule
% \multicolumn{2}{c}{\textbf{Local Spatial Heterogeneity Descriptor}} \\
\textbf{Group} & \textbf{Feature} \\
\midrule
\multirow{4}{*}{Local Entropy~\cite{martinelli2022athena}}
&Skewness of cells' local Shannon index \\
&Skewness of cells' local Simpson index \\
&Skewness of cells' local max-entropy \\
&Skewness of cells' local richness \\
\midrule
\multirow{8}{*}{Local Interaction score~\cite{martinelli2022athena}}
&Mixing of neoplastic cells in inflammatory cells' region \\
&Mixing of neoplastic cells in connective cells' region \\
&Mixing of neoplastic cells in necrosis cells' region \\
&Mixing of neoplastic cells in non-neoplastic epithelial cells' region \\
&Mixing of inflammatory cells in neoplastic cells' region \\
&Mixing of connective cells in neoplastic cells' region \\
&Mixing of necrosis cells in neoplastic cells' region \\
&Mixing of non-neoplastic epithelial cells in neoplastic cells' region \\
\bottomrule
\end{tabular}
}
\caption{Local Spatial Heterogeneity Descriptors}
\label{athena_local}
\end{table}

For instance, clustered arrangements of different nuclei communities typically result in lower local-level entropy, as the neighboring nuclei are mainly from the same class. Conversely, intermixed arrangements lead to higher local-level entropy due to the diversity of neighboring nuclei classes. However, at the global level, these differing arrangements may yield similar entropy values if the overall count of each nuclei class remains constant, despite their distinct spatial distributions. This highlights the importance of analyzing spatial heterogeneity at both local and global levels to capture the full complexity of cellular arrangements in tissue pathology.

\subsection{Normalization}

In this study, we implemented a two-step normalization process for all handcrafted PathExpert features. The first step addresses potential inaccuracies in nuclei segmentation and classification by HoVer-Net, using a binning operation. Each feature within a patch is assessed based on its percentile relative to other patches in the training split of a WSI task. These percentiles are then categorized into 10 discrete bins, ranging from the 0-10$^{th}$ to the 90-100$^{th}$ percentile, effectively shifting the scale of features from absolute values to a relative range from very-low to very-high. This approach transforms feature values into a robust and interpretable format across different patches. The second step involves mean and standard deviation normalization, once again using the training split data. This step centers the data around zero, optimizing it for effective processing by neural networks.

We emphasize that the normalization process alters the scale of the features. For instance, the skewness properties listed in Table~\ref{athena_local} would typically be near 0, negative, or positive in their absolute scale. However, after mean-standard deviation and binning normalization, the scale of skewness may shift, with a 0 skew potentially appearing on either the positive or negative side. Hence, for a more accurate interpretation of our predictions in the local slide-level interpretable predictions (refer to Figure~\ref{fig:patient_interpretable}), it is crucial to consider this scaling effect. Readers should interpret the features as being generally in the lower or higher range and then conceptually approximate these back to their absolute scale. This approach ensures a more nuanced understanding of the predictions post-normalization.

\subsection{Feature Visualization} 

\begin{figure*}[ht!]
\centering
    \includegraphics[width=0.8\linewidth]{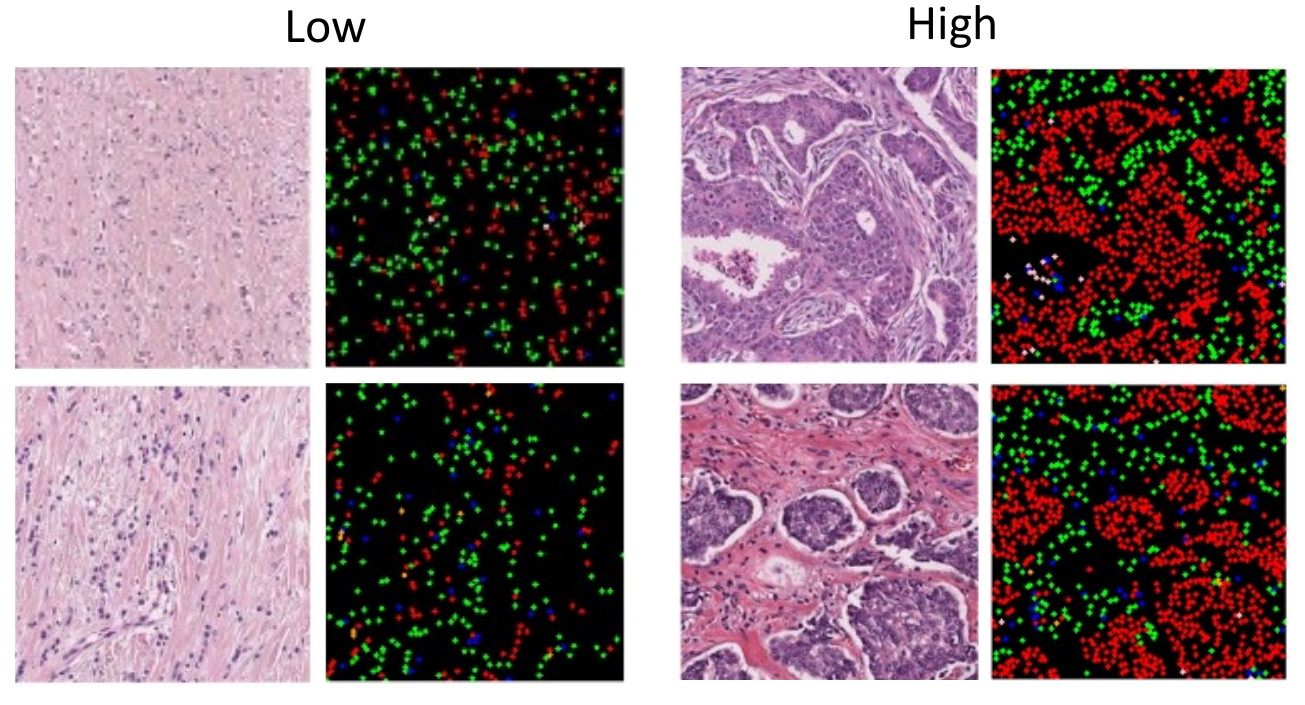}
    \caption{\textbf{Graph Modularity with Cell Types as Community.} As illustrated, in the patches under the column named `Low', cancer cells (in red) co-occur in close spatial proximity with other cell types such as connective (in green) and inflammatory cells (in blue). This results in interconnections among different cell classes when constructing a graph for this feature, leading to low graph modularity. Whereas on the right side, under the column named `High', cells of different classes/communities are more distinctly separated and grouped, resulting in more connections within the same community in the k-nearest neighbor graph, leading to higher graph modularity. In histopathology, this feature can serve as a proxy for distinguishing ductal versus single file line patterns in IDC versus ILC classification in TCGA-BRCA.}
    \label{graph modularity}
\end{figure*}

\begin{figure*}[ht!]
\centering
    \includegraphics[width=0.8\linewidth]{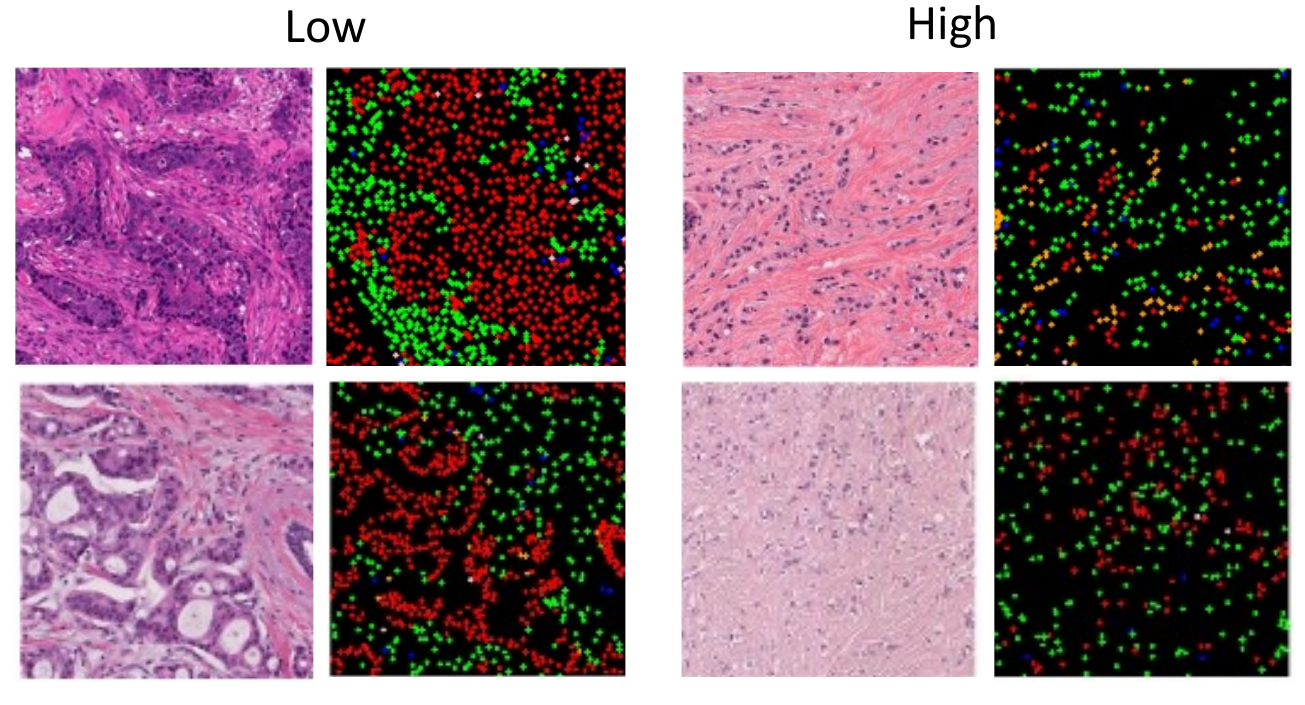}
    \caption{\textbf{Infiltration of Connective Cells in Neoplastic Cells' Region.} In contrast to the scenario presented in Figure~\ref{graph modularity}, the `High' column patches display cancer cells (colored in red) closely intermingled with connective cells (colored in green). This proximity results in more interactions between these cell types in the graph-based analysis of this feature, leading to a marked increase in the infiltration of connective cells within the neoplastic area. On the other hand, in the `Low' column, there is a clearer segregation and clustering of the two cell classes, manifesting in reduced connectivity between them in the k-nearest neighbor graph, and consequently, lower levels of infiltration. In histopathological analysis, this characteristic can be instrumental in differentiating ductal cancers, which show minimal infiltration by other cell types, from invasive patterns characterized by a significant presence of connective cells within the neoplastic areas. This explanation is also applicable to features like the Infiltration of Inflammatory Cells in Neoplastic Cells' Region.
}
    \label{Infiltration of Connective Cells in Neoplastic Region}
\end{figure*}

In the following figures (Figures \ref{Neoplastic Cells: Mean of Their Eccentricity} to \ref{Infiltration of Connective Cells in Neoplastic Region}), we present a visual exploration of some representative features by showcasing patches with low and high values of these features. 
Each figure is accompanied by a detailed caption that elucidates the feature in the context of the patches, providing insights into what constitutes low and high values with respect to that specific feature. 
This visual representation aids in understanding the impact of these features on the tissue's pathology and offers a deeper perspective on how they manifest in different patches.

Note that the terms `cell' and `nucleus' are used interchangeably. However, since the imaging modality is H\&E, all the features actually pertain to nuclei.

% morphometric

\begin{figure*}[ht!]
\centering
    \includegraphics[width=0.75\linewidth]{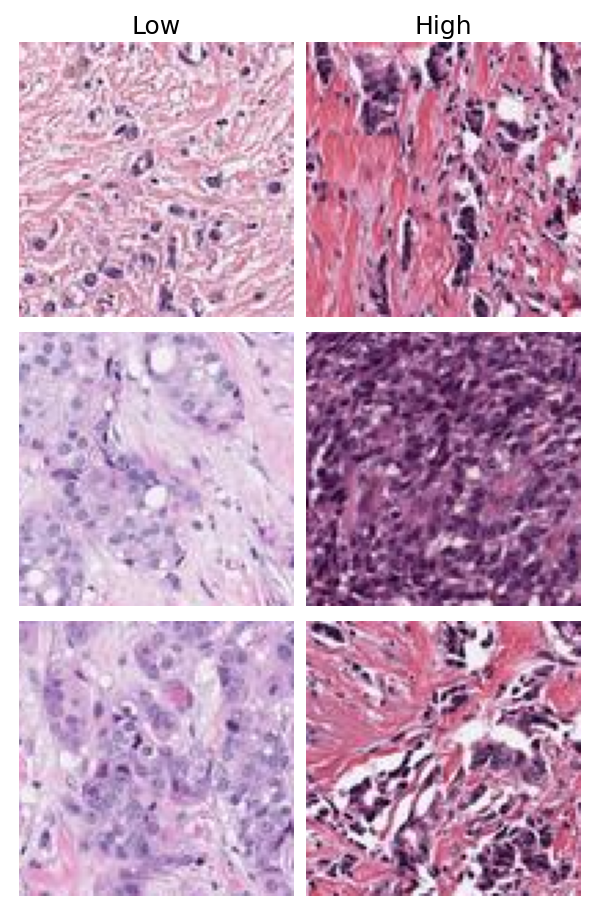}
    \caption{\textbf{Neoplastic Cells: Mean of Eccentricity.} As illustrated, in the patches under the column named `Low', there are round cancer cells (the larger ones), whereas on the right side, under the column named `High', elliptical cells are present, indicating a higher mean of eccentricity. In histopathology, this feature refers to the average deviation of cancer cells from a perfect circular shape. A higher mean eccentricity, as observed in the `High' column, suggests more elliptical cells, often associated with more aggressive or advanced cancer forms.
}
    \label{Neoplastic Cells: Mean of Their Eccentricity}
\end{figure*}

% \begin{figure*}[ht!]
% \centering
%     \includegraphics[width=0.8\linewidth]{feature_visualization_figures/low_medium_high_range_neoplastic cells_ mean of their area.png}
%     \caption{\textbf{Neoplastic Cells: Mean of Area}}
% \end{figure*}

\begin{figure*}[ht!]
\centering
    \includegraphics[width=0.75\linewidth]{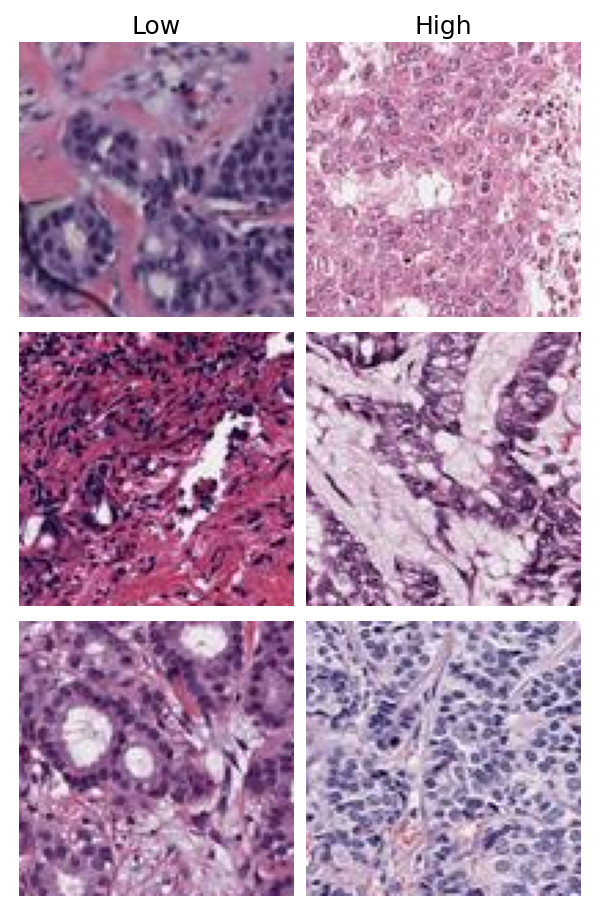}
    \caption{\textbf{Neoplastic Cells: Mean of Intensity Standard Deviation.} As illustrated, in the patches under the column named `Low', there are cancer cells (the larger ones) with uniform intensity, thus the standard deviation is low for each cell, leading to a low mean intensity standard deviation. Whereas on the right side, under the column named `High', the cells exhibit anisochromasia, indicating a higher mean intensity standard deviation. In histopathology, this feature refers to the average deviation of cancer cells from homogeneous intensity. A higher value of this feature, as observed in the `High' column, suggests more anisochromasia, often associated with more aggressive or advanced cancer forms.
}
\end{figure*}

% social network

% \begin{figure*}[ht!]
% \centering
%     \includegraphics[width=0.8\linewidth]{feature_visualization_figures/low_medium_high_range_mean of cells_clustering_coefficient.png}
%     \caption{\textbf{Mean of Cells' Clustering Coefficient}}
% \end{figure*}

\begin{figure*}[ht!]
\centering
    \includegraphics[width=0.72\linewidth]{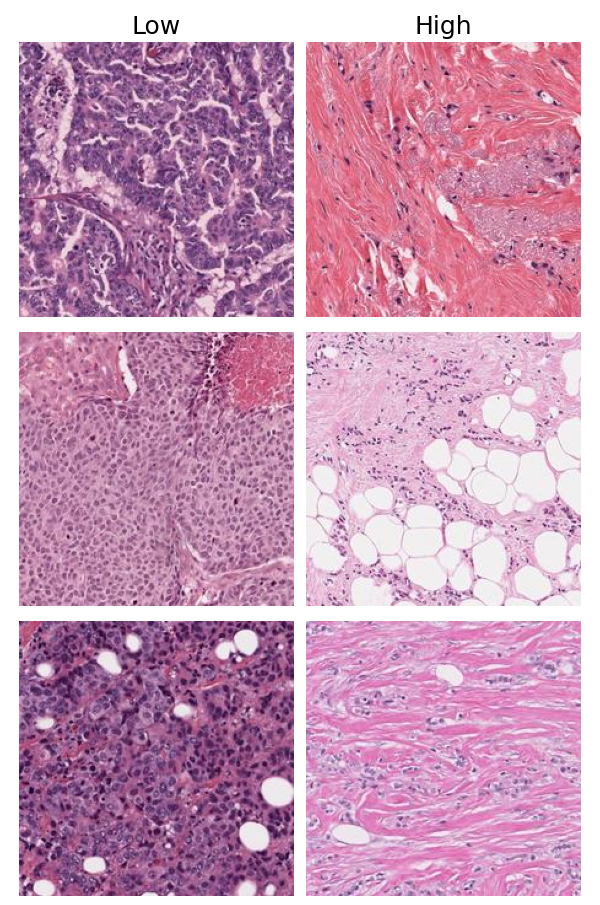}
    \caption{\textbf{Standard Deviation of Cells' Degree.} As illustrated, in the patches under the column named `Low', there is a homogeneous distribution of cells of all types. For this feature, we first construct a k-nearest neighbor graph from cells' centroid and then calculate the cell's degree for each cell. Therefore, a homogeneous distribution leads to each cell having a similar degree, resulting in a lower value of standard deviation. Whereas on the right side, under the column named `High', the cells are much more randomly distributed (disorganized), with grouping in some areas and sparse cells in others. This leads to some cells having a higher degree and others lower, resulting in a high standard deviation of cells' degree in a patch. In histopathology, this feature loosely refers to cohesive versus non-cohesive or homogeneous versus heterogeneous distribution in a spatial context.}
\end{figure*}

% heterogeneity

% \begin{figure*}[ht!]
% \centering
%     \includegraphics[width=0.8\linewidth]{feature_visualization_figures/low_medium_high_range_Gloabl Shannon index.png}
%     \caption{\textbf{Global Shannon Index}}
% \end{figure*}

% \begin{figure*}[ht!]
% \centering
%     \includegraphics[width=0.8\linewidth]{feature_visualization_figures/low_medium_high_range_local shannon index skew.png}
%     \caption{\textbf{Local Shannon Index Skew}}
% \end{figure*}

\end{document}